%% file: main.tex
\documentclass[10pt,twocolumn,letterpaper]{article}

\usepackage[pagenumbers]{cvpr} 

\input{preamble}

\definecolor{cvprblue}{rgb}{0.21,0.49,0.74}
\usepackage[pagebackref,breaklinks,colorlinks,citecolor=cvprblue]{hyperref}

\def\paperID{256} 
\def\confName{3DV\xspace}
\def\confYear{2025\xspace}

\newcommand{\papername}{Spann3R}

\title{3D Reconstruction with Spatial Memory
\vspace{-0.3cm}
}

\author{Hengyi Wang \quad Lourdes Agapito\\
Department of Computer Science, University College London\\
{\tt\small \{hengyi.wang.21, l.agapito\}@ucl.ac.uk}
}
\begin{document}
\twocolumn[{%
    \renewcommand\twocolumn[1][]{#1}%
    \maketitle
    \centering
    \vspace{-0.6cm}
    \input{Figures/teaser}
    \vspace{0.6cm}
}]
\input{sec/0_abstract}    
\input{sec/1_intro}

\input{sec/2_related}

\input{sec/3_method}

\input{sec/4_exp}

\input{sec/5_concl}
\input{sec/X_suppl}

{
    \small
    \bibliographystyle{ieeenat_fullname}
    \bibliography{main}
}

\end{document}

%% file: preamble.tex
\usepackage[dvipsnames]{xcolor}

\usepackage{graphicx}
\usepackage{gensymb}
\usepackage{multirow}
\usepackage{array}
\usepackage{makecell}
\usepackage{tabularx}
\usepackage{booktabs}

\usepackage{amsmath}
\usepackage{amssymb}

\usepackage{url}

\usepackage[accsupp]{axessibility} 

\def\gF{{\mathcal{F}}}

\def\gL{{\mathcal{L}}}

\def\gV{{\mathcal{V}}}

\def\sR{{\mathbb{R}}}

%% file: Figures/teaser.tex
\includegraphics[width=\linewidth]{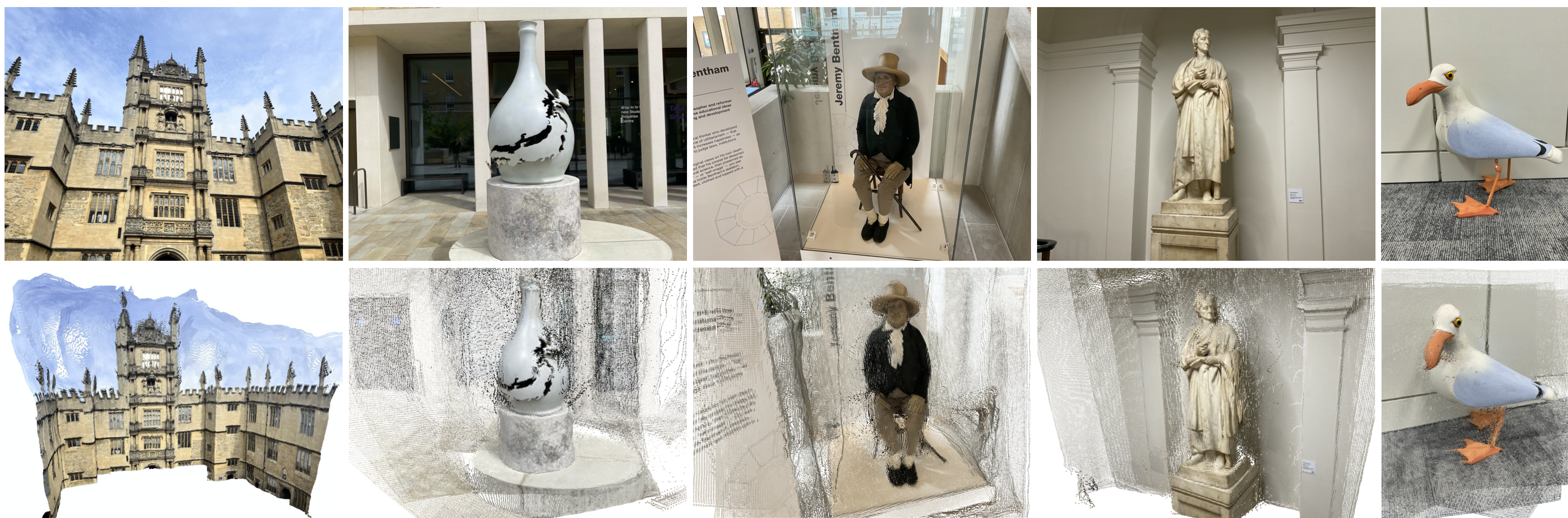}
\captionof{figure}{\textbf{Overview.} Given a set of ordered or unordered image collections without prior knowledge of camera parameters, the proposed \papername{} can incrementally reconstruct the 3D geometry by directly regressing the pointmap of each image in a common coordinate system. \papername{} does not require any optimization-based alignment during inference, i.e., the 3D reconstruction of each image can be solved by a simple forward pass with a transformer-based architecture, thus enabling online reconstruction in real-time. The qualitative examples shown are reconstructed from some self-captured images to illustrate the generalization ability of \papername{}.}

%% file: sec/0_abstract.tex
\begin{abstract}
We present \papername{}, a novel approach for dense 3D reconstruction from ordered or unordered image collections. 
Built on the DUSt3R paradigm, \papername{} uses a transformer-based architecture to directly regress pointmaps from images without any prior knowledge of the scene or camera parameters. 
Unlike DUSt3R, which predicts per image-pair pointmaps each expressed in its local coordinate frame, \papername{} can predict per-image pointmaps expressed in a global coordinate system, thus eliminating the need for optimization-based global alignment.
The key idea of \papername{} is to manage an external spatial memory that learns to keep track of all previous relevant 3D information. \papername{} then queries this spatial memory to predict the 3D structure of the next frame in a global coordinate system. 
%
Taking advantage of DUSt3R's pre-trained weights, and further fine-tuning on a subset of datasets, \papername{} shows competitive performance and generalization ability on various unseen datasets and can process ordered image collections in real-time. Project page: \href{https://hengyiwang.github.io/projects/spanner}{https://hengyiwang.github.io/projects/spanner} 


%

\end{abstract}

%% file: sec/1_intro.tex
\section{Introduction}
\label{sec:intro}

\input{Figures/mot}

Reconstructing dense geometry from images is one of the fundamental problems in computer vision that has been researched for decades~\cite{hartley2003multiple}. This task offers numerous applications in autonomous driving, virtual reality, robotics, medical imaging, and more. The inherent ambiguities in interpreting 3D structures have led traditional solutions evolve into various sub-fields, including keypoint detection and matching~\cite{lowe1999object,lowe2004distinctive,bay2008speeded,rublee2011orb}, Structure-from-Motion (SfM)~\cite{crandall2011discrete,wilson2014robust,sweeney2015optimizing,snavely2006photo,agarwal2009building,wu2013towards,schonberger2016colmap}, Bundle Adjustment (BA)~\cite{triggs2000bundle,agarwal2010bundle,wu2011multicore}, Multi-View Stereo (MVS)~\cite{furukawa2009furu,schonberger2016pixelwise,galliani2015massively}, Simultaneous Localization and Mapping (SLAM)~\cite{davison2007monoslam,klein2007parallel,newcombe2011dtam}, etc. 
Each of these sub-fields addresses different aspects of the problem using a variety of handcrafted heuristics, requiring substantial engineering effort to integrate them into a complete dense reconstruction pipeline~\cite{schonberger2016colmap,schonberger2016pixelwise}.

Recent attention has shifted towards replacing handcrafted features with learned structural priors from large-scale datasets~\cite{detone2018superpoint,sarlin2020superglue,yao2018mvsnet,bloesch2018codeslam,sun2021loftr,yin2023metric3d,wang2024vggsfm,he2024dfsfm}. 
These modern approaches typically integrate learning-based models into each step of the traditional pipeline.
Thus, the sequential structure of traditional pipelines, involving matching, triangulation, sparse reconstruction, camera parameter estimation, and dense reconstruction is mostly maintained.
While these methods have made significant progress with learned priors, the inherent limitations of this complex pipeline persist, making it sensitive to noise at each step and still demanding substantial engineering effort for integration.

To address these issues, DUSt3R~\cite{wang2024dust3r} introduces a radical and novel paradigm shift that was often considered impossible - directly regressing the pointmap, a common representation in visual localization~\cite{brachmann2017dsac,brachmann2023accelerated,revaud2024sacreg,brachmann2024scene}, from a pair of images without prior scene information. Since the pointmap is expressed in the local coordinate system of the image pair,  a global alignment is introduced for the reconstruction of more than just an image pair. This involves per-scene optimization to align the predicted pointmap with a dense pairwise graph.  Trained on millions of image pairs with ground-truth annotations for depth and camera parameters, DUSt3R~\cite{wang2024dust3r} shows unprecedented performance and generalization across various real-world scenarios with different camera sensors. However, operating on a pair of images and the need for per-scene optimization-based global alignment limit its ability for real-time incremental reconstruction and scalability to many images.

In this paper, we present Spann3R, a framework that adopts a \underline{Spa}tial \underline{M}emory for \underline{3}D \underline{R}econstruction. Building on the paradigm of DUSt3R~\cite{wang2024dust3r}, we take a step further by eliminating the need for per-scene optimization-based alignment (See Fig.~\ref{fig:motivation}). That being said, our model enables incremental reconstruction by predicting the pointmap of each image in a common coordinate system with a simple forward pass on our transformer-based architecture. The key idea is to maintain an external memory that keeps track of previous states and learns to query all relevant information from this memory for predicting the next frame, a concept often referred to as memory networks~\cite{weston2014memory,sukhbaatar2015end, miller2016key}.

We employ a lightweight transformer-based memory encoder to encode previous predictions as memory values. To retrieve information from this memory, we project geometric features from two decoders into query features and memory keys using two multilayer perceptron (MLP) heads. 
Our model is trained on sequences of five frames randomly sampled from videos, with a curriculum training strategy that adjusts the sample window size throughout the training process. This allows \papername{} to learn both short and long-term dependency across frames. 
During inference, we apply a memory management strategy inspired by X-Mem~\cite{cheng2022xmem}, which mimics human memory model~\cite{atkinson1968human}, to maintain a compact memory representation. 
Compared to DUSt3R~\cite{wang2024dust3r}, our method aligns point on-the-fly (like a spanner) purely based on neural network (NN), enables real-time online incremental reconstruction at over 50 frames per second (fps) without test-time optimization.  Experiments on various unseen datasets show competitive dense reconstruction quality and generalization ability.

%% file: Figures/mot.tex
\begin{figure*}[t]
    \centering
    \includegraphics[width=\textwidth]{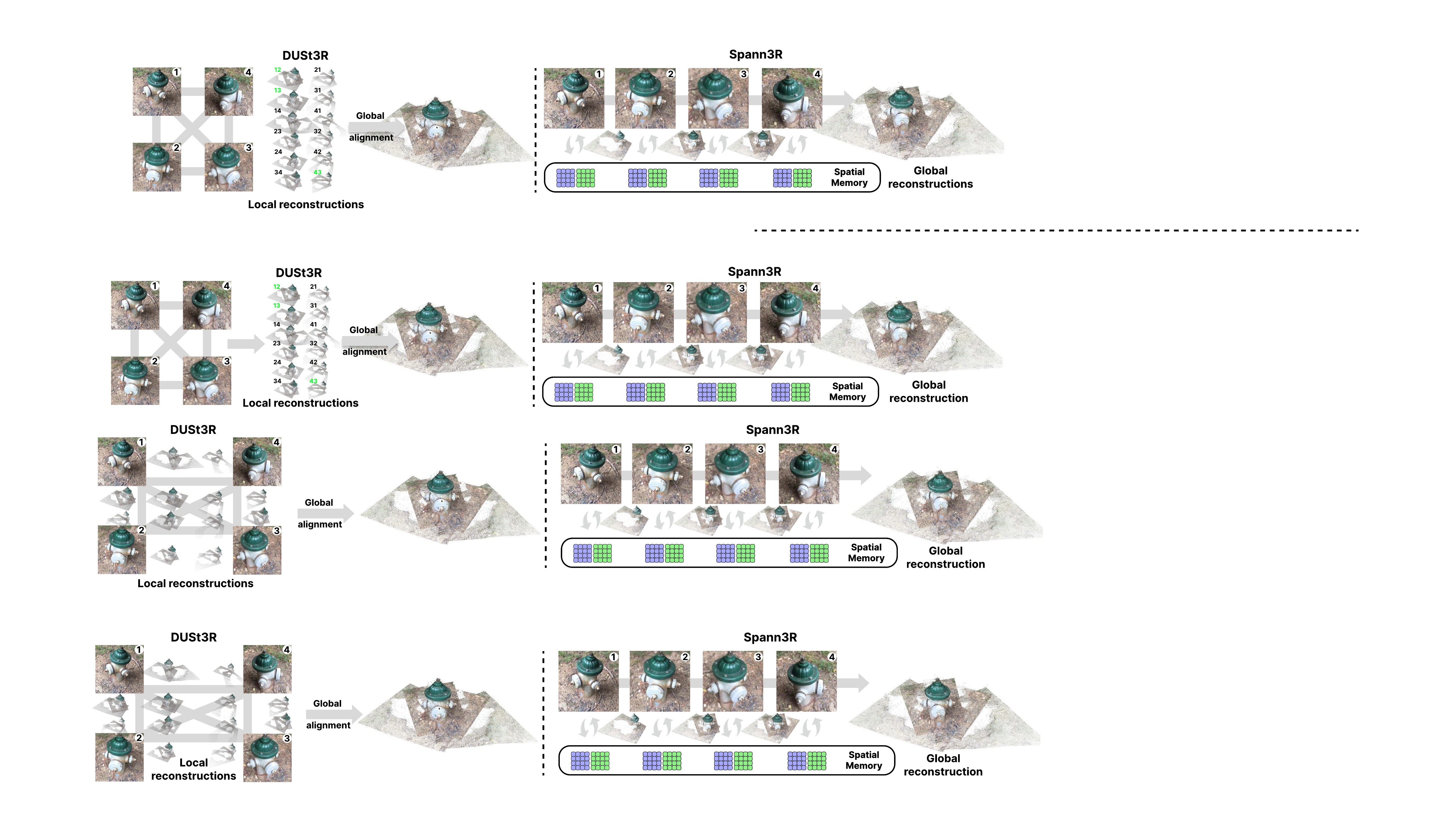}
    \vspace{-10pt}
    \caption{\textbf{Motivation.} DUSt3R~\cite{wang2024dust3r} directly regresses the pointmap of each image pair in a local coordinate system. In contrast,  \papername{} predicts a global pointmap in a common coordinate system via a spatial memory that stores all previous predictions. Thus, our method can enable online incremental reconstruction without the need to build a dense pairwise graph and a final optimization-based alignment.}
    \label{fig:motivation}
    \vspace{-5pt}
\end{figure*}

%% file: sec/2_related.tex
\section{Related Works}

\input{Figures/method}

\noindent
\textbf{Classic 3D Reconstruction.} Recovering 3d structures from visual signals has been investigated for decades~\cite{hartley2003multiple}. Structure-from-motion (SfM)~\cite{crandall2011discrete,wilson2014robust,sweeney2015optimizing,snavely2006photo,agarwal2009building,wu2013towards,schonberger2016colmap,pan2024glomap} is usually considered the de-facto standard for obtaining sparse geometry and accurate camera poses. Starting from feature correspondence search ( keypoint detection and description~\cite{lowe1999object,lowe2004distinctive,bay2008speeded,rublee2011orb}, matching~\cite{agarwal2009building,wu2013towards}, and geometric verification~\cite{schonberger2016colmap}), SfM selects an image pair for initialization, followed by image registration, triangulation, and bundle adjustment~\cite{triggs2000bundle,agarwal2010bundle,wu2011multicore}. Finally, multi-view stereo~\cite{furukawa2009furu,schonberger2016pixelwise,galliani2015massively} is used to obtain dense geometry. These approaches usually require lengthy offline optimization. In contrast, visual SLAM focuses on obtaining geometry online in real-time. Given the calibrated cameras, visual SLAM can perform sparse~\cite{davison2007monoslam,klein2007parallel,mur2015orb,engel2017direct} or dense~\cite{newcombe2011dtam,engel2014lsd} reconstruction via minimizing either reprojection error (indirect)~\cite{davison2007monoslam,klein2007parallel,mur2015orb} or photometric error (direct)~\cite{newcombe2011dtam,engel2014lsd,engel2017direct}. To obtain accurate reconstruction, these methods either require a depth/LiDAR sensor~\cite{newcombe2011kinectfusion} or careful initialization and various assumptions about the camera motion and scene appearance~\cite{davison2007monoslam,mur2015orb,newcombe2011dtam}.

\noindent
\textbf{Learning-based 3D Reconstruction.} Built upon the success of the classic reconstruction pipeline, recent approaches usually leverage learning-based techniques to improve each sub-task, i.e., feature extraction~\cite{detone2018superpoint,yi2016lift}, matching~\cite{sarlin2020superglue,lindenberger2023lightglue}, BA~\cite{lindenberger2021pixel}, monocular depth estimation~\cite{dexheimer2023learning,yin2023metric3d,ke2024repurposing}, multi-view depth estimation~\cite{yao2018mvsnet,duzceker2021deepvideomvs,sayed2022simplerecon}, optical flow~\cite{teed2020raft}, point tracking~\cite{doersch2023tapir,karaev2023cotracker,xiao2024spatialtracker}, etc. However, those classic pipelines usually involve a sequential structure vulnerable to noise in each sub-task. To avoid this, DUSt3R~\cite{wang2024dust3r} unifies all sub-tasks by directly learning to map an image pair to 3D, followed by an optimization-based global alignment to bring all image pairs into a common coordinate system. In this work, we take a step further to replace the optimization step with an end-to-end learning framework, enabling online incremental reconstruction in real time.

\noindent
\textbf{Neural Rendering for 3D Reconstruction.} The recent progress in differentiable rendering, i.e., NeRF~\cite{mildenhall2020nerf} and its follow-up works~\cite{barron2022mip,kerbl2023gaussian,barron2023zip,wang2021neus,yariv2021volsdf,jang2021codenerf,jang2024nvist,huang20242d} enables high-fidelity scene reconstruction using images with known camera parameters obtained via SfM~\cite{schonberger2016colmap}. Several other works leverage neural rendering for SfM~\cite{bian2023nope} and SLAM~\cite{sucar2021imap,zhu2022nice,wang2023co,kong2023vmap}. However, despite the significant progress in accelerating neural rendering, these methods still require lengthy optimization time. For instance, Gaussian splatting~\cite{kerbl2023gaussian} and its variants in SLAM~\cite{matsuki2024monogs,keetha2024splatam,huang2024photo} can achieve over 100fps rendering. However, they still require minutes of test-time optimization for scene reconstruction.

\noindent
\textbf{Memory Networks.} The concept of the memory networks was originally introduced in the context of question-answering~\cite{weston2014memory,sukhbaatar2015end, miller2016key} in natural language processing, where they manage an external memory for reasoning over long-term dependencies. This architecture is naturally suitable for processing sequential data and can thus be adopted in various vision tasks, such as video object segmentation (VOS)~\cite{cheng2021stcn,oh2019video,cheng2022xmem,cheng2024putting}, video understanding~\cite{song2024moviechat}, etc. Our work is greatly inspired by a series of works in VOS, where STM~\cite{oh2019video} firstly employs the memory networks for VOS, and XMem~\cite{cheng2022xmem} further extends this idea to long video sequence via a memory consolidation strategy that mimics the human memory model~\cite{atkinson1968human}.

%% file: Figures/method.tex
\begin{figure*}[t]
    \centering
    \includegraphics[width=\textwidth]{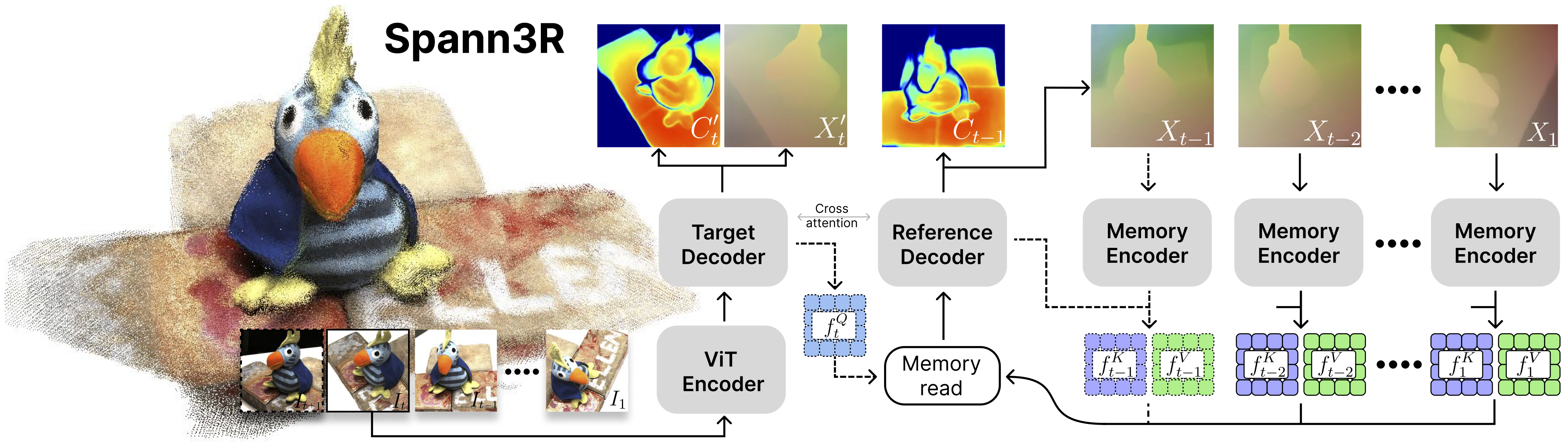}
     \vspace{-5pt}
    \caption{\textbf{Overview of \papername{}.} Our model contains a ViT~\cite{dosovitskiy2020image} encoder and two intertwined decoders as in DUSt3R~\cite{wang2024dust3r}. The target decoder here is used to obtain query features from images for memory query while the reference decoder is used to predict based on the memory readout using the geometric feature and our memory features. A lightweight memory encoder is used to encode the previously predicted pointmap together with geometric features into our memory key and value features. Dash means operation in the next time step.} 
    \label{fig:overview}
    \vspace{-5pt}
\end{figure*}

%% file: sec/3_method.tex
\section{Method}

Fig.~\ref{fig:overview} shows an overview of \papername{}. We aim to repurpose DUSt3R~\cite{wang2024dust3r} into an end-to-end incremental reconstruction framework that directly regresses the pointmap in a common coordinate system. Specifically, given a sequence of images $\{I_t\}_{t=1}^N$, our goal is to train a network $\gF$ that maps each $I_t$ to its corresponding pointmap $X_t$, expressed in the coordinate system of the initial frame. To enable this, we introduce a spatial memory that encodes the previous predictions for reasoning next frame. We will describe our network architecture (Sec.~\ref{sec:method1}), spatial memory (Sec.~\ref{sec:method2}), and the training and inference of our model (Sec.~\ref{sec:method3}) next.

\subsection{Network architecture}
\label{sec:method1}

\noindent
\textbf{Feature encoding.} In each forward pass, our model takes a frame $I_t$ and a previous query $f_{t-1}^Q$ as input. A ViT~\cite{dosovitskiy2020image} is used to encode the frame $I_t$ into visual feature $f_t^I$:
\begin{equation}
    f_t^I = \mathrm{Encoder}^I(I_t).
\end{equation}

\noindent
The query features $f_{t-1}^Q$ is used to retrieve features in our memory bank to output the fused feature $f_{t-1}^G$:
\begin{equation}
    f_{t-1}^G = \mathrm{Memory\_read}(f_{t-1}^Q, f^K, f^V),
\end{equation}

\noindent
where $f^K$ and $f^V$ are memory key and value features.

\noindent
\textbf{Feature decoding.} The fused feature $f_{t-1}^G$ and the visual feature $f_t^I$ are fed into two intertwined decoders that process them jointly via cross-attention. This can enable the model to reason the spatial relationship between two features:
\begin{equation}
    f_t^{H^\prime}, f_{t-1}^H = \mathrm{Decoder}(f_t^I, f_{t-1}^G).
\end{equation}
The feature $f_t^{H^\prime}$ decoded by the target decoder is fed into an MLP head to generate the query feature for the next step:
\begin{equation}
    f_{t}^Q = \mathrm{head}_{\mathrm{query}}^{\mathrm{target}}(f_t^{H^\prime}, f_t^I). \label{eq:query}
\end{equation}
\noindent
The feature $f_{t-1}^H$ decoded by the reference decoder is fed into an MLP head to generate the pointmap and confidence:
\begin{equation}
    X_{t-1}, C_{t-1} = \mathrm{head}_{\mathrm{out}}^{\mathrm{ref}}(f_{t-1}^H).
 \end{equation}

\noindent
Note that we also generate a pointmap and confidence from $f_t^{H^\prime}$ only for supervision. 

\noindent
\textbf{Memory encoding}. The feature and predicted pointmap of the reference decoder are used for encoding the memory key and value features:
\begin{align}
     f_{t-1}^K &= \mathrm{head}_{\mathrm{key}}^{\mathrm{ref}}(f_{t-1}^H, f_{t-1}^I),\\
     \label{eq:mem_k}
      f_{t-1}^V &= \mathrm{Encoder}^V(X_{t-1}) + f_{t-1}^K.
\end{align}

\noindent
Since memory key and value features have information from both geometric features and visual features, it enables memory readout based on both appearance and distance.

\noindent
\textbf{Discussion.} Compared to DUSt3R~\cite{wang2024dust3r}, our architecture has one more lightweight memory encoder and two MLP heads for encoding the query, memory key, and memory value features. For decoders, DUSt3R~\cite{wang2024dust3r} contains two decoders - a reference decoder that reconstructs the first image in the canonical coordinate system, and a target decoder that reconstructs the second image in the coordinate system of the first image. In contrast, we repurpose the two decoders in DUSt3R~\cite{wang2024dust3r}. The target decoder is mainly used to produce features for querying the memory while the reference decoder takes the fused features from memory for reconstruction. In terms of the initialization, we directly use two visual features.

\subsection{Spatial memory}
\label{sec:method2}

Fig.~\ref{fig:sp_memory} shows an overview of the spatial memory that consists of a dense working memory, a sparse long-term memory, and a memory query mechanism for extracting features from the memory, which we will describe next.

\noindent
\textbf{Memory query.} The spatial memory stores all key $f^K\in\sR^{Bs\times (T\cdot P)\times C}$ and value  $f^V\in\sR^{Bs\times (T\cdot P)\times C}$ features. To compute fused feature $f_{t-1}^G$, we apply a cross attention using query feature $f_{t-1}^Q\in\sR^{Bs\times P\times C}$ for memory reading:
\begin{equation}
    f_{t-1}^G = A_{t-1} f^V + f_{t-1}^Q,
\end{equation}

\noindent
where $A_{t-1}\in\sR^{Bs\times P\times (T\cdot P)}$ is the attention map:
\begin{equation}
    A_{t-1} = \mathrm{Softmax}(\frac{f_{t-1}^Q (f^K)^\top}{\sqrt{C}}).
    \label{eq:attn_weight}
\end{equation}

\noindent 
This attention map contains a dense attention weight for each token in the current query with respect to all tokens in memory keys (See Fig.~\ref{fig:attn}). We apply an attention dropout of $0.15$ during training to encourage the model to reason the geometry from a subset of the memory values. 

In practice, we observe that at inference, most of the attention weights are relatively small, as illustrated in the cumulative histogram of Fig.~\ref{fig:sp_memory}. However, despite their small weights, the corresponding patches can be significantly distant from the query patches or even outliers. In the end, their memory values might still have a non-negligible impact on the fused features. To mitigate the impact of these outlier features, we apply a hard clipping threshold of $5 \times 10^{-4}$ and re-normalize the attention weights.

\noindent
\textbf{Working memory.} The working memory consists of dense memory features from the most recent 5 frames. For each incoming memory feature, we first correlate its key feature with each key feature in working memory. We only insert new key and value features into working memory if their maximum similarity is less than $0.95$. Once the working memory is full, the oldest memory features are drained into long-term memory.

\noindent
\textbf{Long-term memory.} During the inference, long-term memory features accumulate over time, which can increase GPU memory usage and slow down speed. Inspired by XMem~\cite{cheng2022xmem}, which mimics human memory models~\cite{atkinson1968human} via memory consolidation, we design a similar strategy to sparsify the long-term memory. Specifically, for each token in long-term memory keys, we keep track of its accumulated attention weights (i.e., $A$ in Eq.~\ref{eq:attn_weight}). Once the long-term memory reaches a predefined threshold, we perform memory sparsification by retaining only the top-k tokens.

\subsection{Training and Inference}
\label{sec:method3}

\input{Figures/method_memory}

\input{Figures/qual}

\noindent
\textbf{Objective function.} Following Dust3R~\cite{wang2024dust3r}, we train our model by a simple confidence-aware regression loss. We additionally include a scale loss to encourage the average distance of the predicted point cloud to become smaller than the ground truth. The overall loss is

\begin{equation}
    \mathcal{L} =  \mathcal{L}_{\mathrm{conf}} + \mathcal{L}_{\mathrm{scale}}.
\end{equation}

\noindent
Note that to compute $\mathcal{L}_{\mathrm{conf}}$, both the predicted and ground truth pointmaps are normalized by their average distance. We tried to fix this scale based on the first two-view prediction during initial experiments, but it does not work well due to the presence of outliers and the unbounded nature of the outdoor scene, Co3D~\cite{reizenstein2021common}, for instance.

\noindent
\textbf{Curriculum training.} Due to GPU memory constraints, we train our model by randomly sampling 5 frames per video sequence. Thus, the memory bank contains only a 4-frame  memory at maximum during training. To ensure the model adapts to diverse camera motions and long-term feature matching, we gradually increase the sample window size throughout the training. For the last 25\% epochs, we gradually decrease the window size to ensure the training frame interval aligns with the inference frame interval.

\noindent
\textbf{Inference.} Our model naturally fits sequential data, i.e. video sequence. For unordered image collections, we can build a dense pairwise graph as in DUSt3R~\cite{wang2024dust3r}. The pair with the highest confidence will be used for initialization. Then, we can either build a minimum spanning tree based on pairwise confidence to determine the order or directly feed the remaining images into our model to identify the next best image based on the predicted confidence. Note that the confidence map in DUSt3R~\cite{wang2024dust3r} involves an exponential function, which tends to overweight patches with higher confidence. In our case, we find that map it back to a sigmoid function for view selection can improve the robustness of the reconstruction.

%% file: Figures/method_memory.tex
\begin{figure}[t]
    \centering
    \includegraphics[width=0.99\columnwidth]{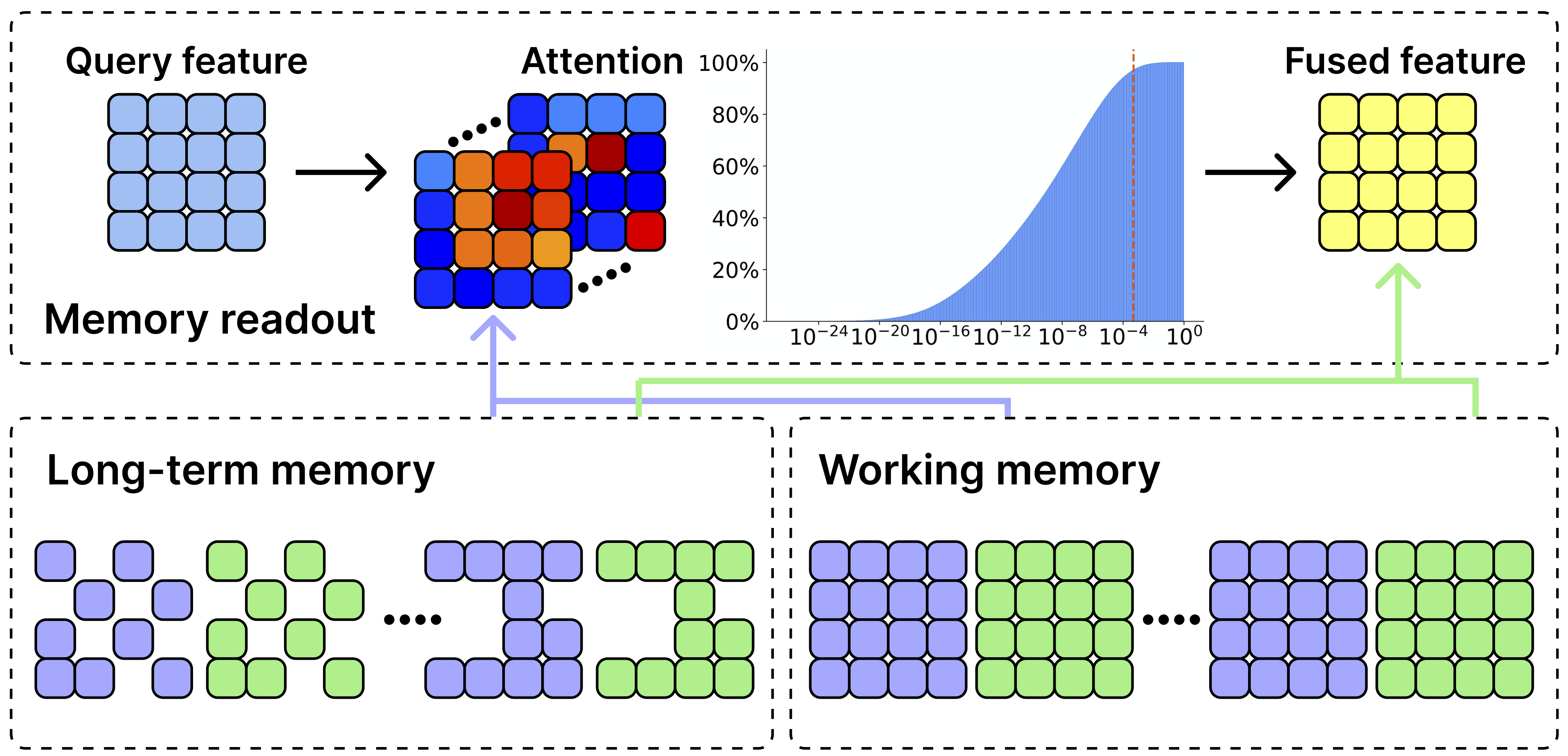}
    \caption{\textbf{Overview of our spatial memory.} Our memory contains a dense working memory chunk and a sparse long-term memory chunk. For each memory query, all tokens in both long-term memory and working memory will be used for generating attention weight and the fused feature. We also visualize the cumulative histogram of the values in attention weight. }
    \label{fig:sp_memory}
    \vspace{-5pt}
\end{figure}

%% file: Figures/qual.tex
\begin{figure*}[tp]
  \centering
  \setlength{\tabcolsep}{1.5pt}
  \newcommand{\sz}{0.14}
  \begin{tabular}{lccccc}
    & FrozenRecon~\cite{xu2023frozenrecon} & Dust3R$^\dagger$~\cite{wang2024dust3r} & Ours & GT & Reference \\
    \makecell{\rotatebox{90}{\tt fire-04}} &
    \makecell{\includegraphics[height=\sz\linewidth]{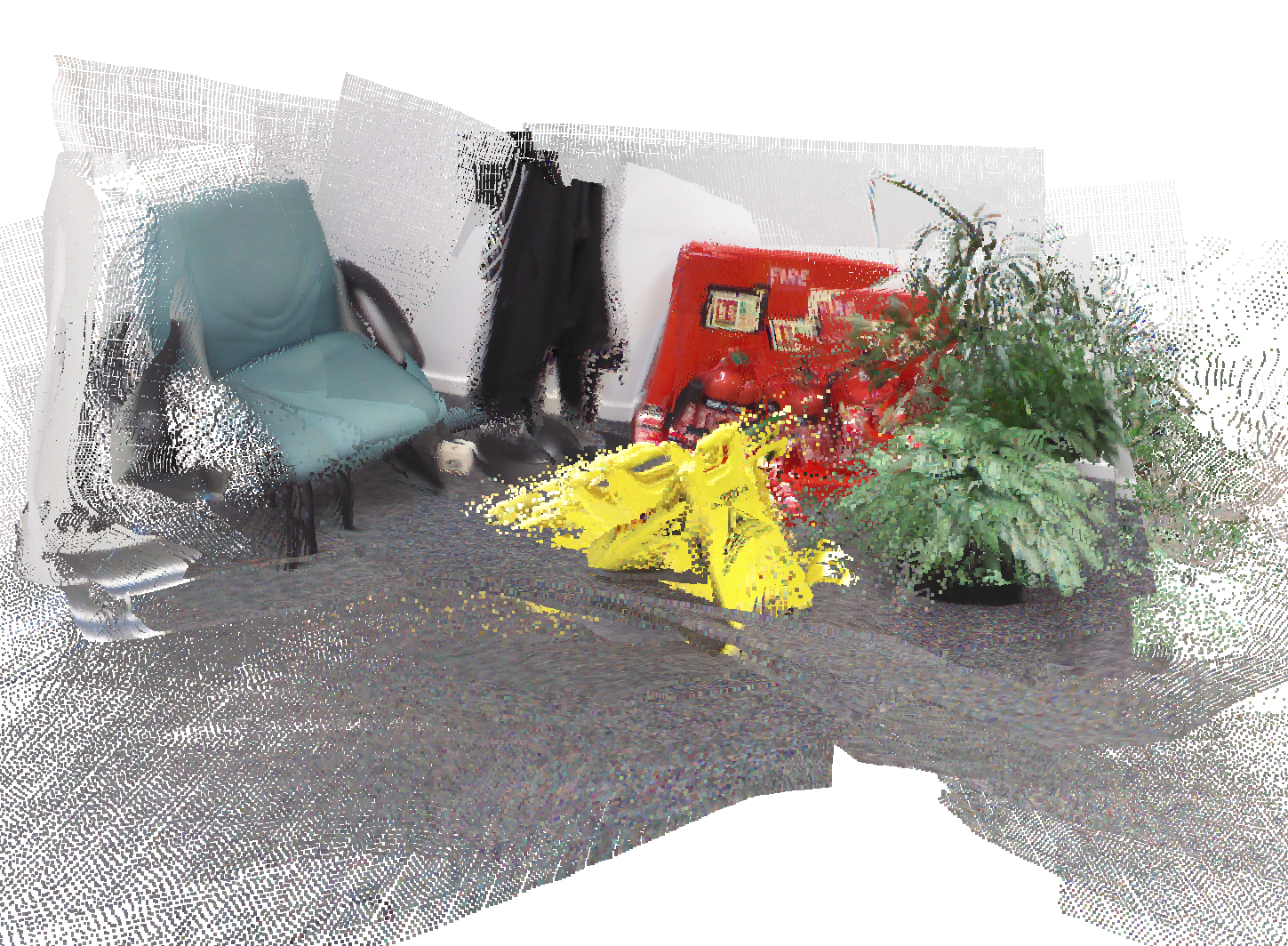}} &
    \makecell{\includegraphics[height=\sz\linewidth]{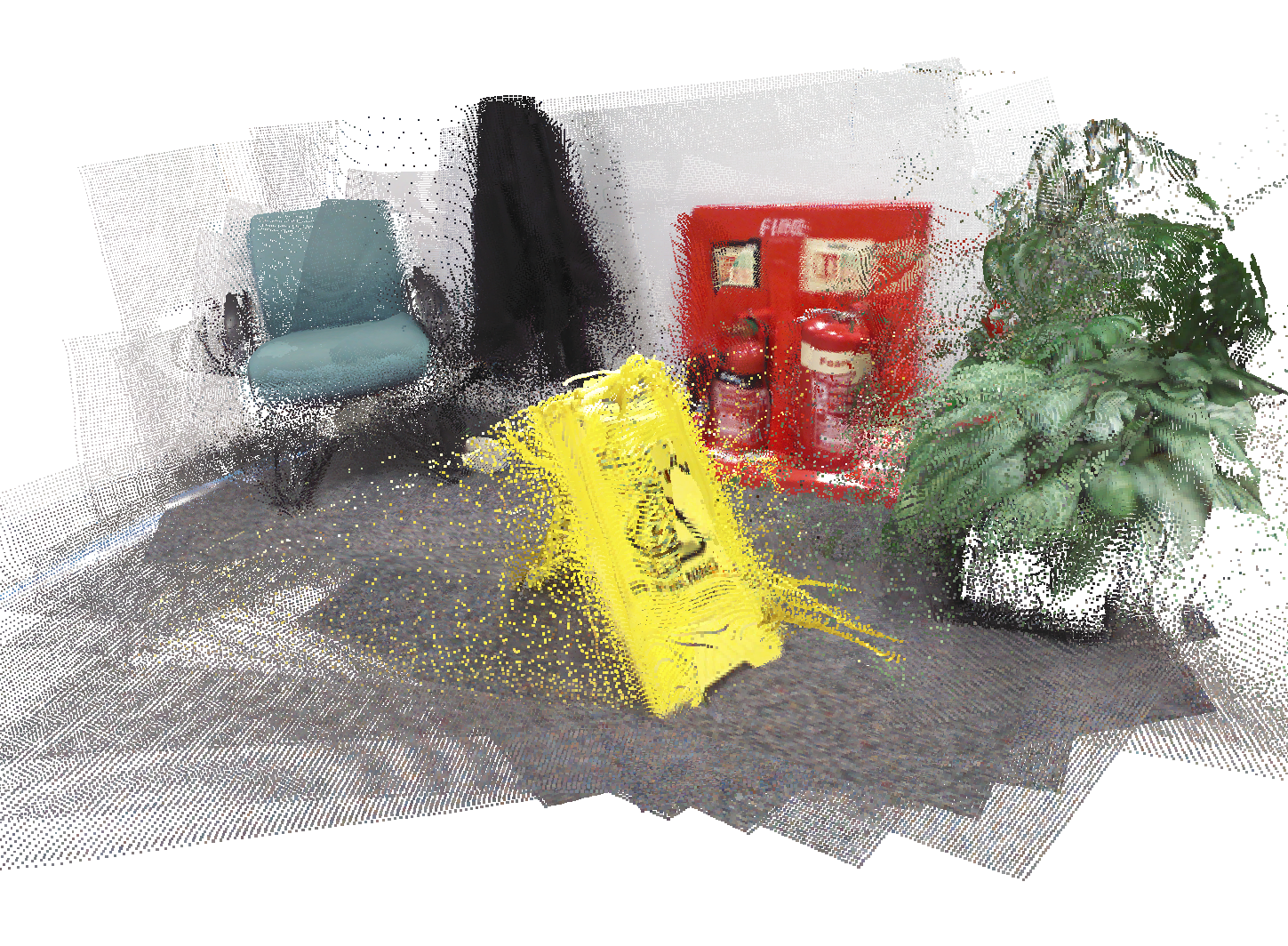}} &
    \makecell{\includegraphics[height=\sz\linewidth]{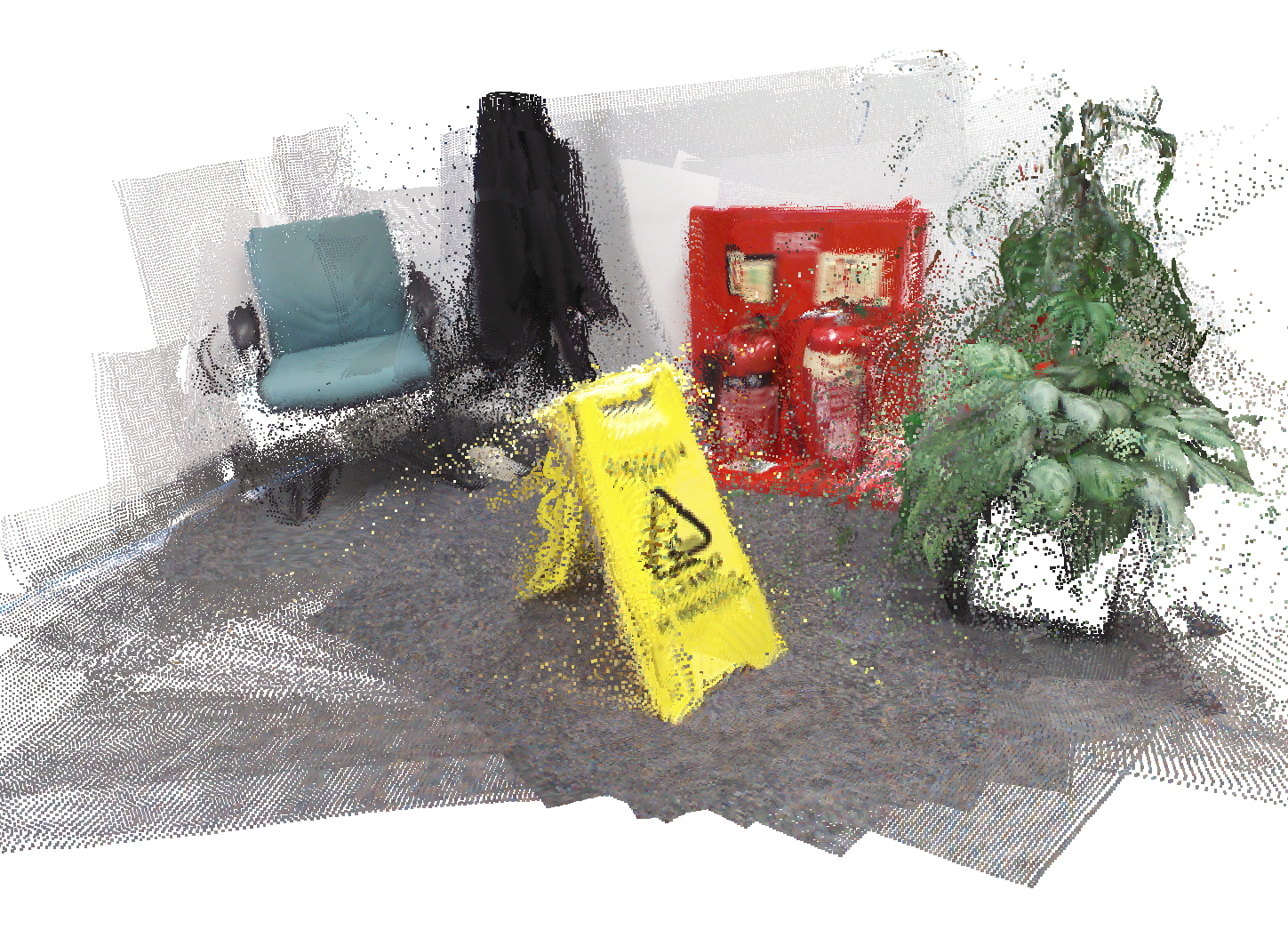}} &
    \makecell{\includegraphics[height=\sz\linewidth]{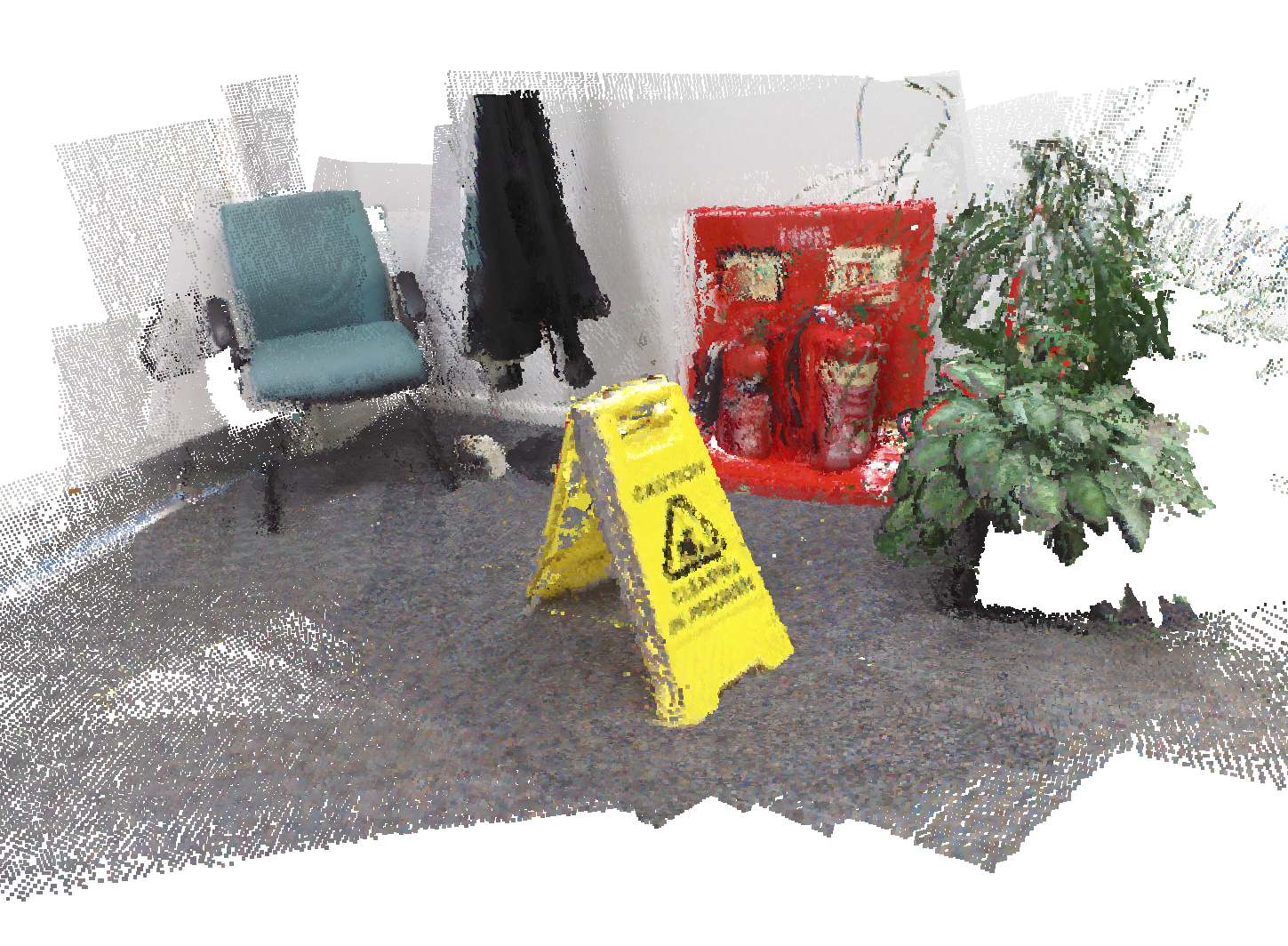}} &
    \makecell{\includegraphics[height=\sz\linewidth]{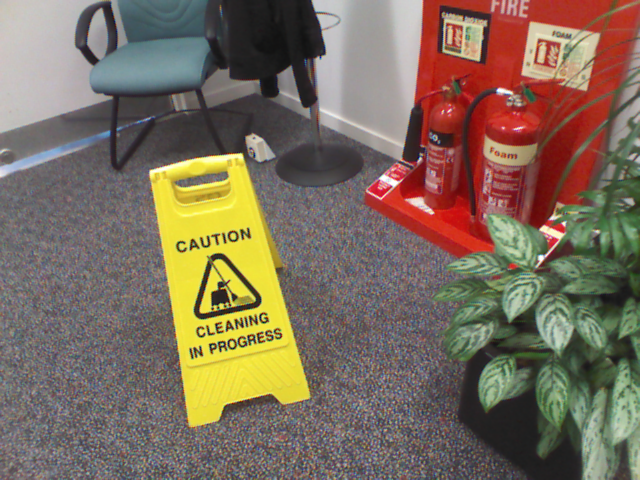}} \\
    \makecell{\rotatebox{90}{\tt redkit-06}} &
    \makecell{\includegraphics[height=\sz\linewidth]{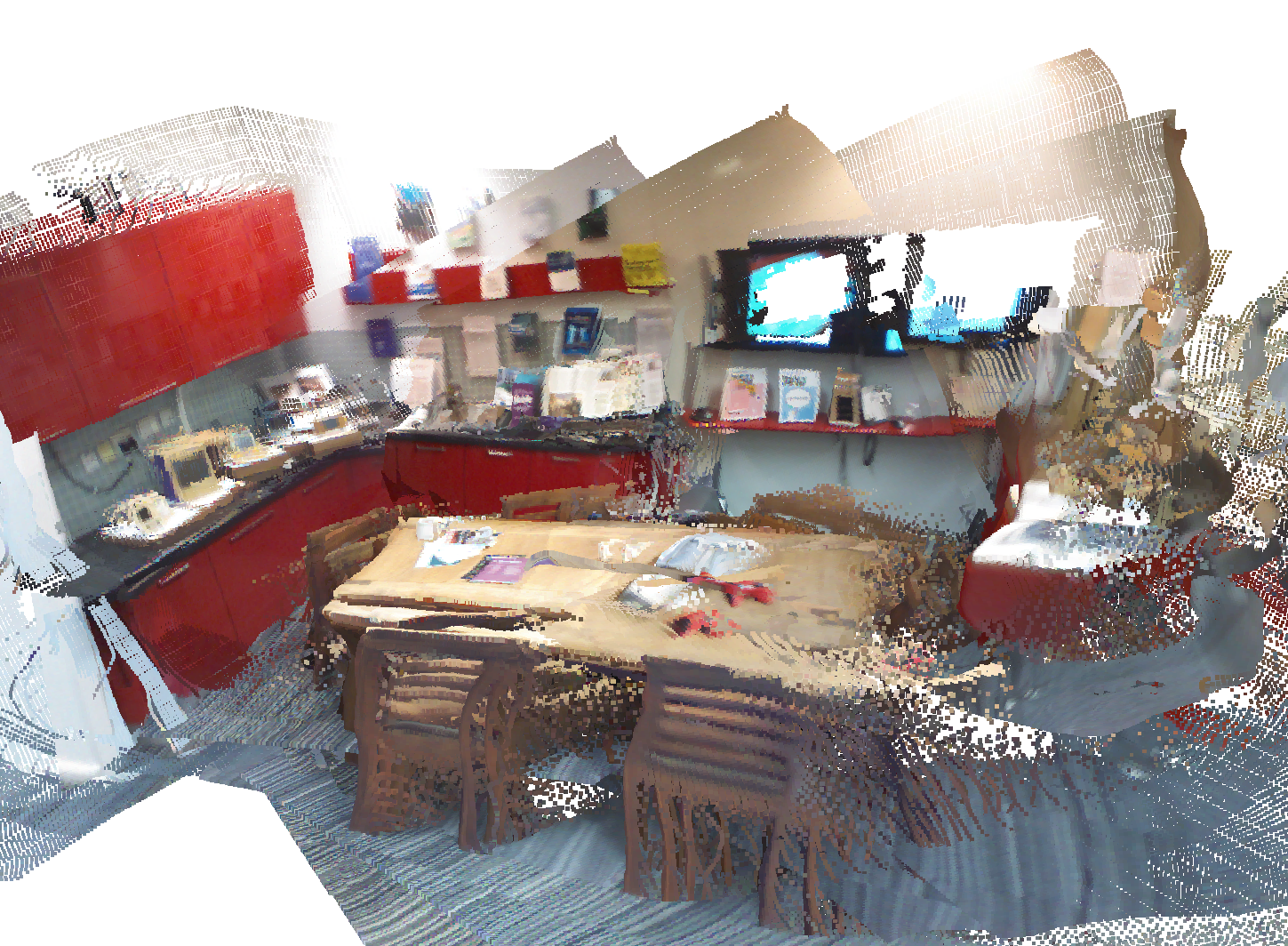}} &
    \makecell{\includegraphics[height=\sz\linewidth]{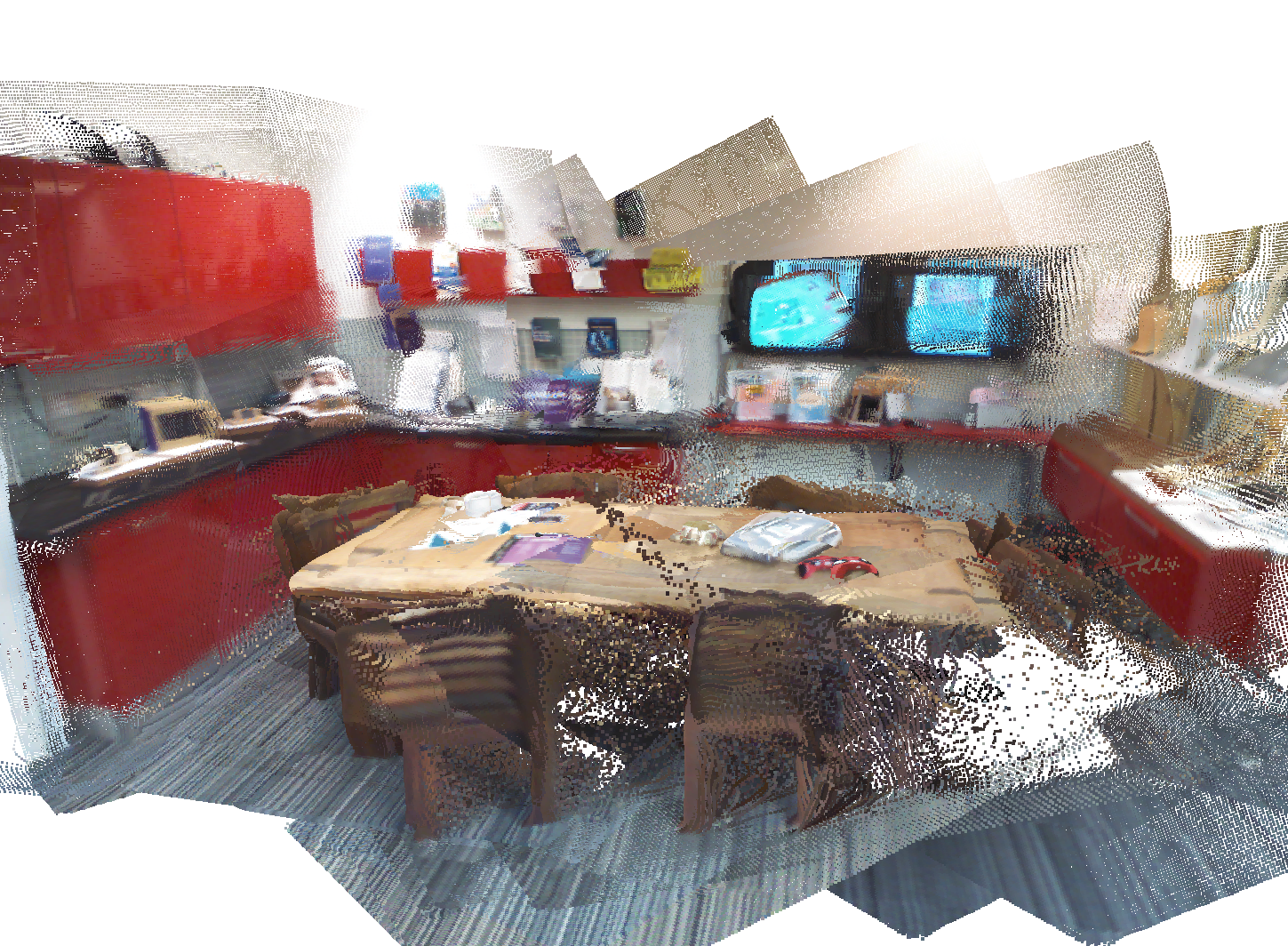}} &
    \makecell{\includegraphics[height=\sz\linewidth]{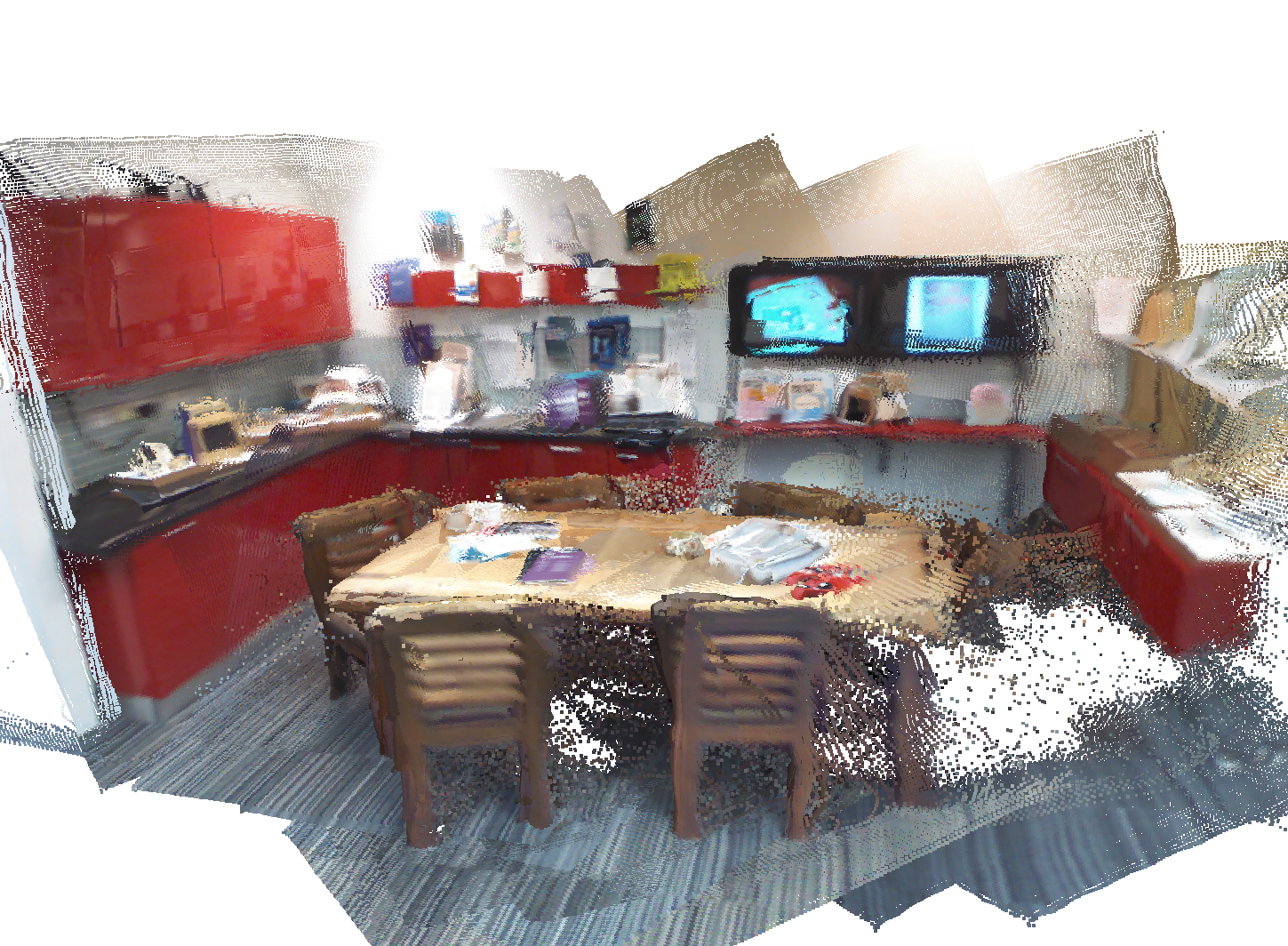}} &
    \makecell{\includegraphics[height=\sz\linewidth]{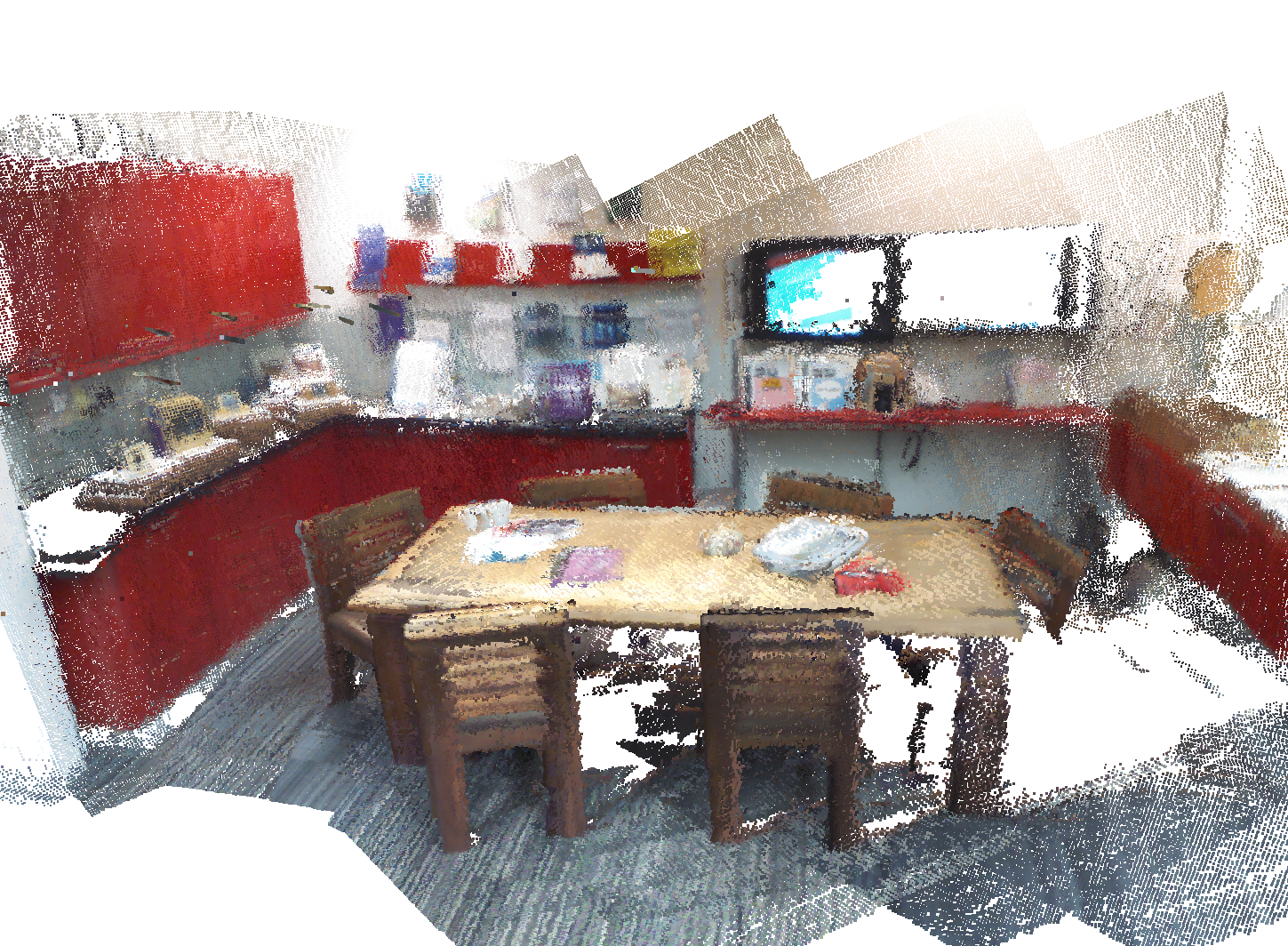}} &
    \makecell{\includegraphics[height=\sz\linewidth]{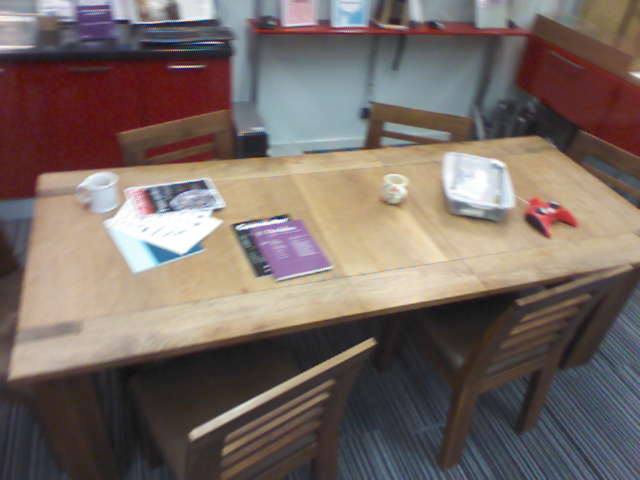}} \\
    \makecell{\rotatebox{90}{\tt office-09}} &
    \makecell{\includegraphics[height=\sz\linewidth]{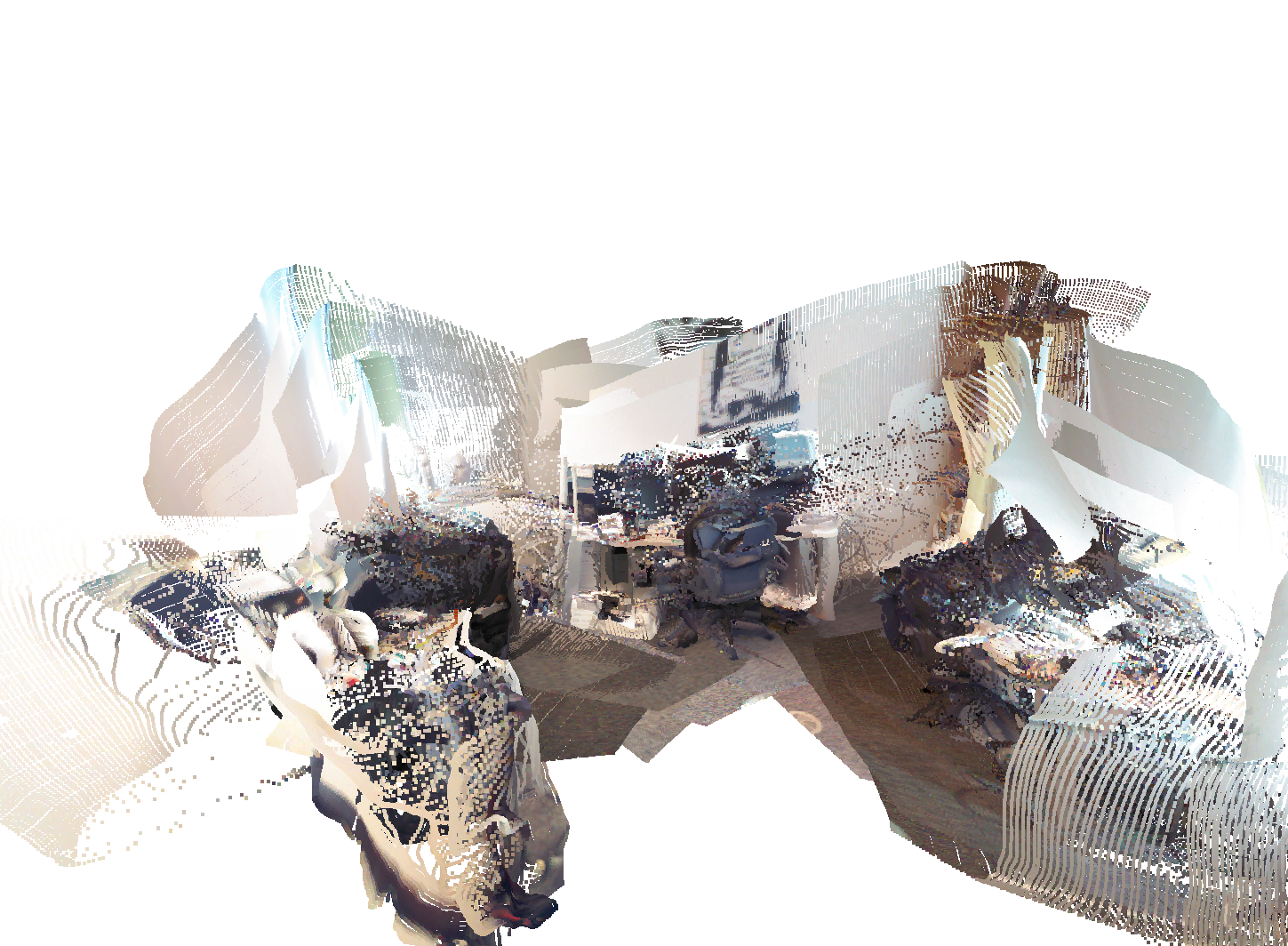}} &
    \makecell{\includegraphics[height=\sz\linewidth]{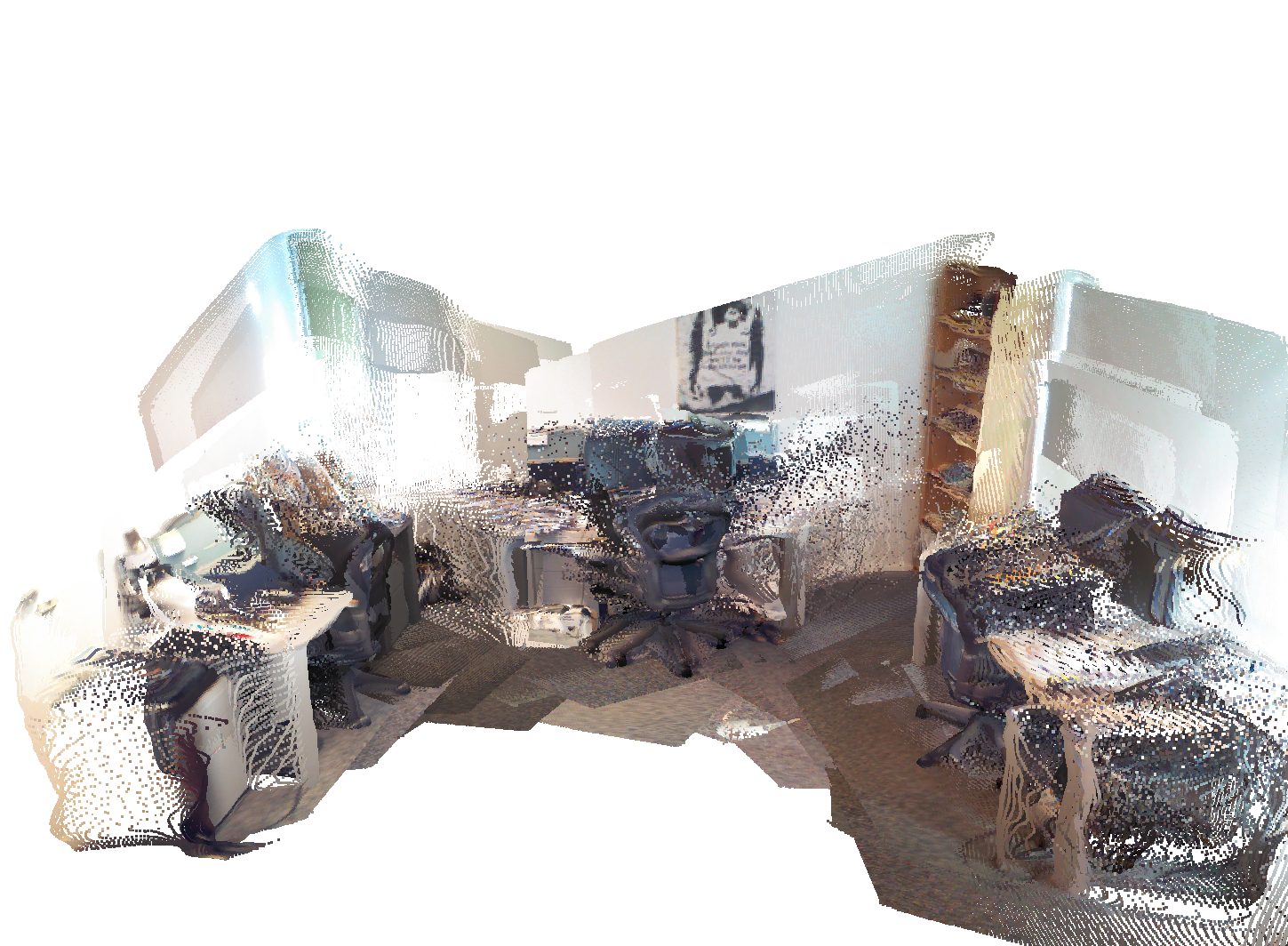}} &
    \makecell{\includegraphics[height=\sz\linewidth]{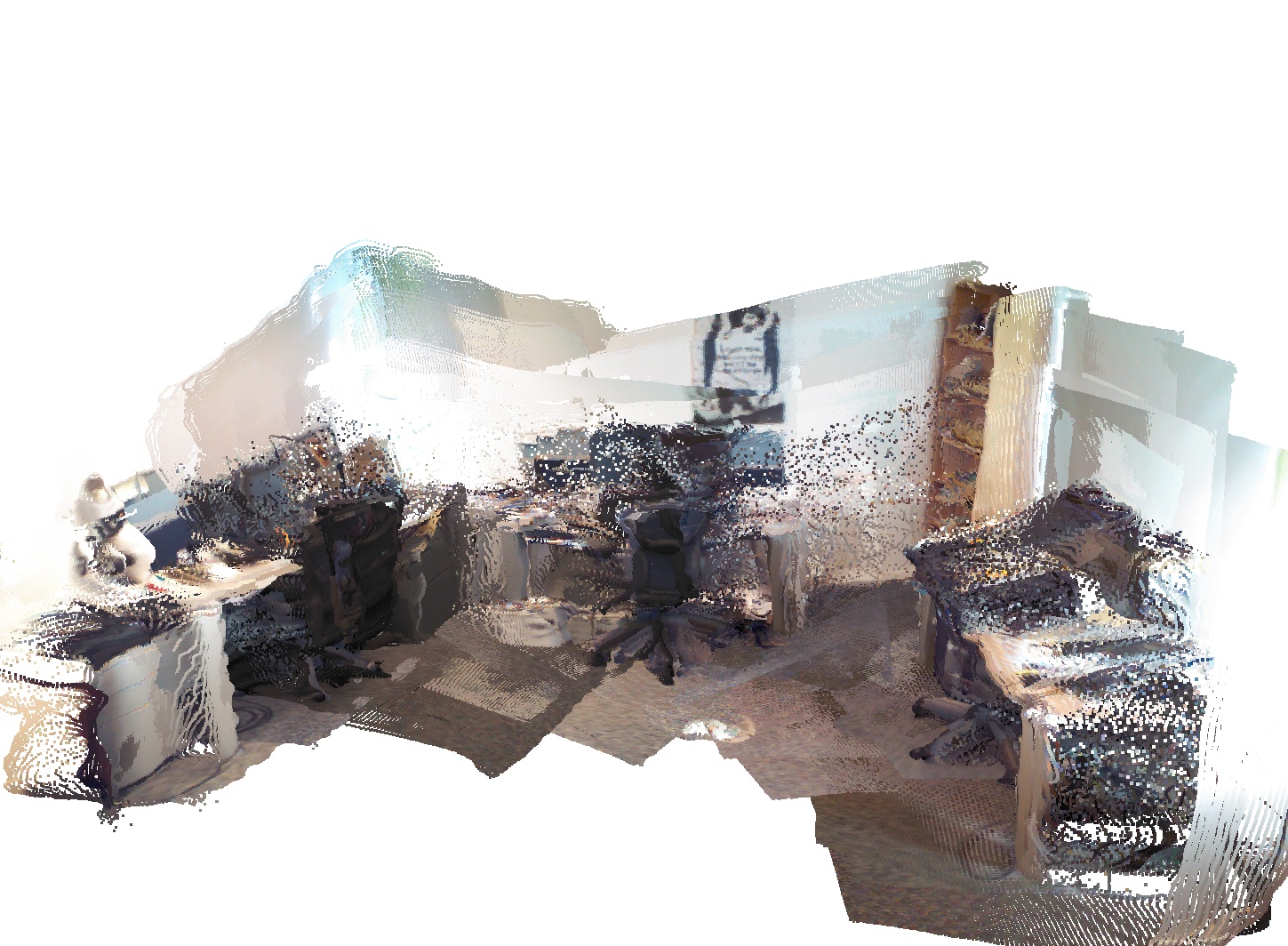}} &
    \makecell{\includegraphics[height=\sz\linewidth]{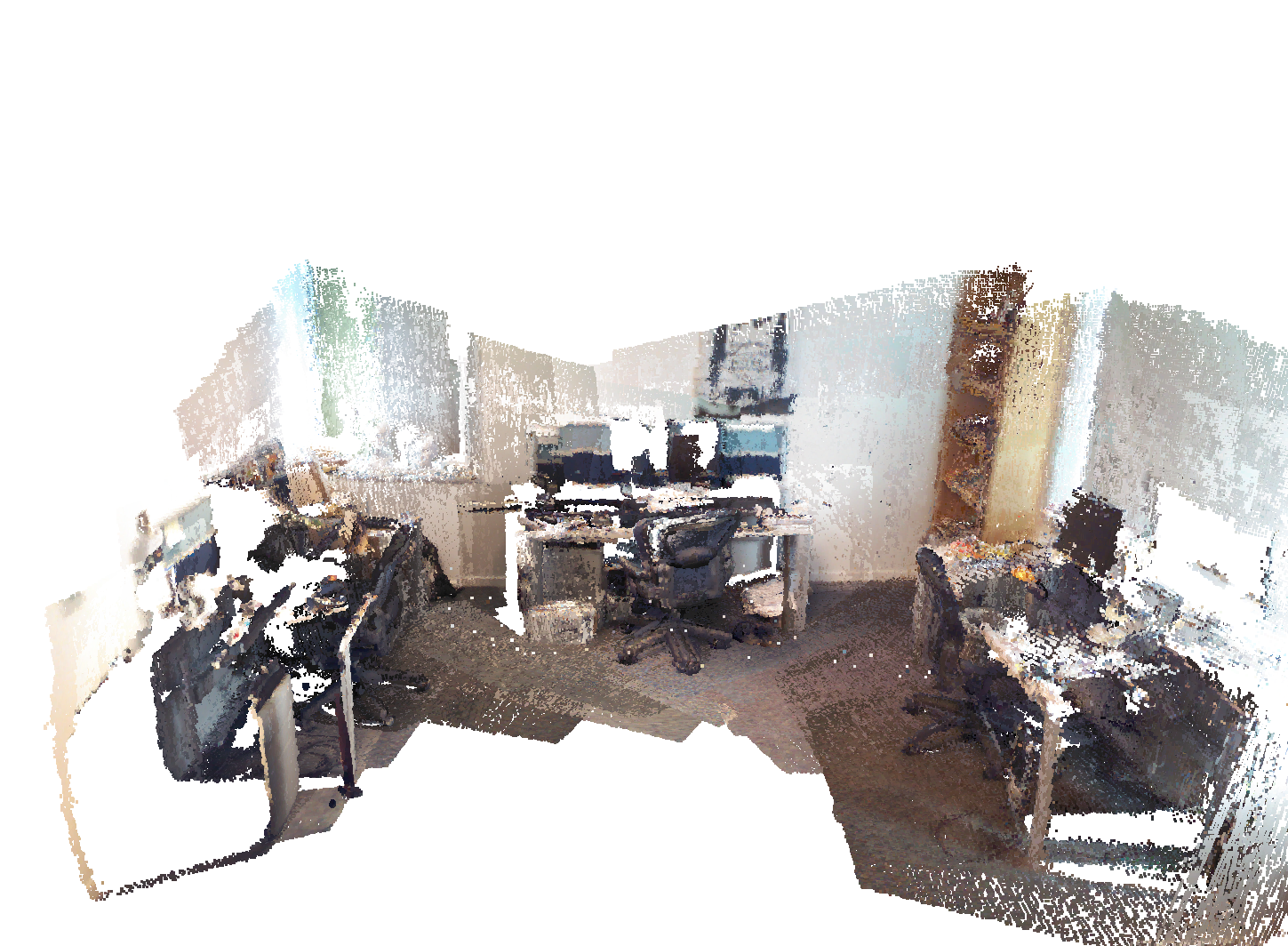}} &
    \makecell{\includegraphics[height=\sz\linewidth]{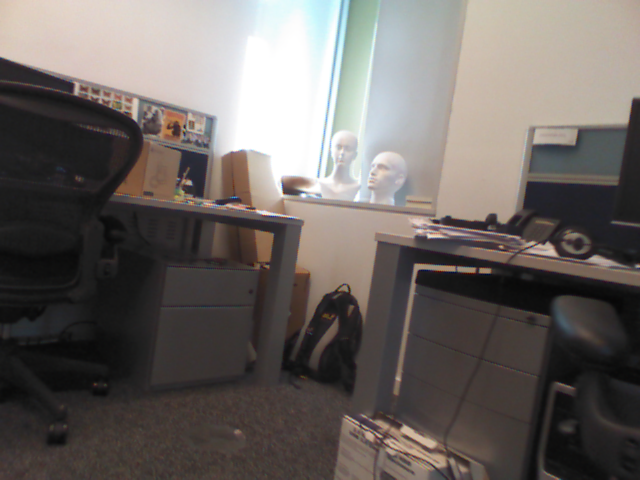}} \\
    
  \end{tabular} 
  \caption{\textbf{Qualitative examples.} We show qualitative examples of DUSt3R$^\dagger$~\cite{wang2024dust3r}, FrozenRecon~\cite{xu2023frozenrecon} for a comprehensive comparison. Our method shows competitive results in comparison to other offline methods. However, since our method runs online without any optimization-based alignment, it can potentially lead to drift issues in some challenging scenarios (See Office-09).}
  \label{fig:qual_comparison}
\end{figure*}

%% file: sec/4_exp.tex
\section{Experiments}

\input{Tables/quant}

\subsection{Setup}
\noindent
\textbf{Datasets.} DUSt3R~\cite{wang2024dust3r} adopts a mixture of 8 datasets: Habitat~\cite{savva2019habitat}, MegaDepth~\cite{li2018megadepth}, ARKitScenes~\cite{baruch1arkitscenes}, Static Scenes 3D~\cite{mayer2016large},  BlendedMVS~\cite{yao2020blendedmvs}, ScanNet++~\cite{yeshwanthliu2023scannetpp}, Co3D-v2~\cite{reizenstein2021common}, and Waymo~\cite{sun2020scalability}. We choose a subset of datasets: Habitat~\cite{savva2019habitat}, ScanNet~\cite{dai2017scannet}, ScanNet++~\cite{yeshwanthliu2023scannetpp}, ARKitScenes~\cite{baruch1arkitscenes}, BlendedMVS~\cite{yao2020blendedmvs}, Co3D-v2~\cite{reizenstein2021common} for training our model. Note that for Habitat~\cite{savva2019habitat}, we only use a small subset of the scenes to synthesize data for training. To demonstrate the generalization ability of our model, we quantitatively evaluate our model on 3 unseen datasets: 7Scenes~\cite{shotton2013scene}, NRGBD~\cite{Azinovic_2022_CVPR}, and DTU~\cite{aanaes2016large}. 

\noindent
\textbf{Baselines.} We consider DUSt3R~\cite{wang2024dust3r} as our primary baseline. Additionally, we compare with FrozenRecon~\cite{xu2023frozenrecon} on indoor scene reconstruction. FrozenRecon is a test-time optimization method that jointly optimizes camera parameters along with the scale and shift factor of the depth map from the off-the-shelf monocular depth estimation model. All evaluations are performed on a single NVIDIA 4090 GPU with 24GB of memory. DUSt3R$^\dagger$ denotes running DUSt3R's final weight with $224\times 224$ images as running full reconstruction on $512\times 384$ images cannot fit in 24GB memory. We include both results of DUSt3R on few-view reconstruction for a comprehensive comparison.

\noindent
\textbf{Metrics.} We use \textit{accuracy}, \textit{completion} and \textit{normal consistency} as in prior works~\cite{Azinovic_2022_CVPR,zhu2022nice,wang2023co}. The predicted dense pointmap is directly compared with the back-projected per-point depth, excluding invalid and background points if applicable. Since the reconstruction is up to an unknown scale, we align the reconstruction following DUSt3R~\cite{wang2024dust3r}. For DUSt3R$^\dagger$ and our method, $224\times 224$ inputs are generated using resizing and center cropping. Evaluation on DUSt3R and FrozenRecon with full-resolution images is restricted to the same $224\times 224$ region for fairness. However, evaluating on full-resolution (4:3) may benefit from increased visual overlapping compared to $224\times 224$ (1:1).

\noindent
\textbf{Implementation details.} We initialize part of our model with pre-trained weights from DUSt3R~\cite{weinzaepfel2023croco,wang2024dust3r} with ViT-large~\cite{dosovitskiy2020image} encoder, ViT-base decoders, and a DPT head~\cite{ranftl2021vision}. For the memory encoder, we employ a light-weight ViT containing 6 self-attention blocks with the embedding dimension of 1024. Due to the computational constraint, we only train our model on $224\times 224$ images for $120$ epochs using AdamW optimizer with a learning rate of $5e-5$ and $\beta= (0.9, 0.95)$. The training takes around 10 days on 8 V100 GPUs, each with 32GB memory. The batch size is $2$ per GPU, which leads to the effective batch size of $16$.

\subsection{Evaluation}

\input{Tables/quant_dtu}

\noindent
\textbf{Scene-level reconstruction.} We compare the reconstruction quality with FrozenRecon~\cite{xu2023frozenrecon} and DUSt3R~\cite{wang2024dust3r}, both of which are offline dense reconstruction methods that involve optimization-based alignment. As shown in Tab.~\ref{tab:quantitative}, our model shows competitive online reconstruction quality compared to the other two offline methods while being significantly faster. This is because our model is able to predict the pointmap in a common coordinate system without the need for test-time optimization. For few-view reconstruction, our model achieves performance on par with DUSt3R$^\dagger$. However, since our model is trained on $224 \times 224$ images, it shows a performance gap compared to DUSt3R which uses $512 \times 384$ images for reconstruction, especially on the NRGBD~\cite{Azinovic_2022_CVPR} dataset, which contains many thin structures. Fig.~\ref{fig:qual_comparison} shows three qualitative examples on the 7scenes~\cite{shotton2013scene} dataset, where our model demonstrates comparable results to DUSt3R$^\dagger$. However, due to the absence of bundle adjustment, our model may drift. This is shown in Office-09, where a strong specular reflection in the corner causes inaccurate prediction, eventually leading to drift.

\noindent
\textbf{Object-level reconstruction.} In Tab.~\ref{tab:quant_dtu}, we evaluate the object-level reconstruction on DTU~\cite{aanaes2016large} dataset. DTU contains a challenging camera trajectory, starting from a top-down view, which makes online reconstruction particularly difficult. For offline reconstruction, our method achieves performance on par with DUSt3R$^\dagger$ in terms of median Acc, Comp, and NC. It is important to note that DTU contains a black background with many thin structures (see Fig.~\ref{fig:attn}). As a result, our model may produce floaters around the edges, which receive significant penalties in terms of mean Acc.

\input{Tables/quant_mem}

\input{Figures/map}

\input{Figures/memory_size}

\input{Figures/affinity}

\input{Figures/qual2}

\noindent
\textbf{Run-time and memory footprint.} Our default setting of online reconstruction can run around $65$fps with $11$GB GPU memory on a single 4090 GPU.

\subsection{Analysis}

\noindent
\textbf{Effect of the memory bank.} We conduct an ablation study on our memory bank in Tab.~\ref{tab:ab_memory}. Without long-term memory (w/o lm), the model tends to drift quickly when relying only on the working memory, which consists of the most recent 5 frames. Additionally, without clipping attention weight, the performance of our model can degrade in certain scenes. This occurs because despite most attention weight values being small (See Fig.~\ref{fig:sp_memory}), the corresponding memory values can still differ significantly, especially when the geometry prediction contains outliers. Filtering out those small attention weights can improve the robustness of our reconstruction pipeline in various challenging scenarios. Fig.~\ref{fig:ab_memory} shows the reconstruction quality with respect to different long-term memory sizes. In practice, we find that 4000 memory tokens are sufficient for most scenes.

\noindent
\textbf{Online reconstruction.} We visualize the online reconstruction of two indoor scenes in Fig.~\ref{fig:map}. From the results, we can see our method can achieve the online reconstruction of the indoor scene even in some challenging scenarios (textureless walls). This shows a certain level of understanding of the regularity presented in indoor scenes, i.e. Manhattan World Assumption. However, one limitation is that due to the accumulated errors and outliers, our model may not fill the geometry correctly when the loop closes (See last column of Fig.~\ref{fig:map}).

\noindent
\textbf{Visualization of affinity map.} In Fig.~\ref{fig:attn}, we visualize the attention weights corresponding to different patches in the query frame throughout the memory frames. To read out from memory, we project visual features and geometry features from two decoders into query features and memory key features as in Eq.~\ref{eq:query} and Eq.~\ref{eq:mem_k}. This can help to distinguish the parts with similar appearance and semantics but in different locations (e.g. the right eye and foot of the toy).

\noindent
\textbf{Generalization to other unseen datasets.} We demonstrate the generalization capability of \papername{} through qualitative examples of reconstructions shown in Fig.~\ref{fig:qual_gen}. These examples include results from the Map-free Reloc~\cite{arnold2022mapfree}, ETH3D~\cite{schops2017etb3d}, MipNeRF-360~\cite{barron2022mip}, NeRF~\cite{mildenhall2020nerf} and TUM-RGBD~\cite{sturm2012tumrgbd} datasets. The results illustrate that \papername{} can generalize to different types of scenes and has a certain level of robustness across various challenging scenarios.

\subsection{Discussion}

Despite showing competitive results across various datasets, our method still has some inherent limitations. We will describe several limitations and potential directions next.

\noindent
\textbf{Large-scale scene reconstruction.} Our model can deal with large-scale object-centric scenes fairly well. However, in cases where the camera continuously moves forward or reconstructs large multi-room scenes, our model might fail. This limitation arises due to the limited memory size during training. Since our training process assumes the camera pose of the first frame is the identity, training on just 5 frames typically spans only a limited spatial region. To address this issue, one approach could be to restart our model every few frames and then align the different fragments using PnP-RANSAC. Alternatively, a more scalable sampling strategy in training or a more structured memory system at inference is needed to overcome this challenge.

\noindent
\textbf{Bundle adjustment.} For an incremental reconstruction pipeline, bundle adjustment is usually of great importance for mitigating error accumulation. In the case of \papername{}, the question would be: Can we learn to update and fuse our memory when incorporating new observations? Alternatively, since the concept of \papername{} is to predict the next frame based on previous predictions, we could potentially integrate traditional bundle adjustment techniques to correct the geometry. The model could then encode this corrected geometry into the spatial memory, leading to more accurate predictions in subsequent frames.

\noindent
\textbf{Training data.} Due to the constraint of the computational resources, we only train our model across 4 datasets using five $224\times 224$ images sampled from the entire sequence. We expect training on the entire datasets of DUSt3R~\cite{wang2024dust3r}, either with more than five images or at a higher $512$ resolution, could further improve the accuracy. Moreover, the current model relies on a substantial amount of posed RGB-D data. It is worth exploring how to effectively learn data-driven prior from casual videos using self-supervised training.

%% file: Tables/quant.tex
\begin{table*}[t]
  \centering
  \footnotesize
  \setlength{\tabcolsep}{0.3em}
    \begin{tabularx}{\textwidth}{r c c >{\centering\arraybackslash}X >{\centering\arraybackslash}X >{\centering\arraybackslash}X >{\centering\arraybackslash}X >{\centering\arraybackslash}X >{\centering\arraybackslash}X >{\centering\arraybackslash}X >{\centering\arraybackslash}X >
    {\centering\arraybackslash}X >{\centering\arraybackslash}X >
    {\centering\arraybackslash}X >
    {\centering\arraybackslash}X >{\centering\arraybackslash}X}
      \toprule

           \multirow{3}{*}{Method} & \multicolumn{1}{c}{\multirow{3}{*}{Optim.}} & \multicolumn{1}{c}{\multirow{3}{*}{Onl.}} & \multicolumn{6}{c}{7 scenes} & \multicolumn{6}{c}{NRGBD} & \multirow{3}{*}{FPS} \\

      \cmidrule(lr){4-9} \cmidrule(lr){10-15}
         & & & \multicolumn{2}{c}{\tt{Acc}$\downarrow$} & \multicolumn{2}{c}{\tt{Comp}$\downarrow$} & \multicolumn{2}{c}{\tt{NC}$\uparrow$} & \multicolumn{2}{c}{\tt{Acc}$\downarrow$} & \multicolumn{2}{c}{\tt{Comp}$\downarrow$} & \multicolumn{2}{c}{\tt{NC}$\uparrow$} & \\
      \cmidrule(lr){4-5} \cmidrule(lr){6-7} \cmidrule(lr){8-9} \cmidrule(lr){10-11} \cmidrule(lr){12-13} \cmidrule(lr){14-15}
         & & & \tt{Mean} & \tt{Med.} & \tt{Mean} & \tt{Med.} & \tt{Mean} & \tt{Med.} & \tt{Mean} & \tt{Med.} & \tt{Mean} & \tt{Med.} & \tt{Mean} & \tt{Med.} & \\
      \midrule
      \textbf{F-Recon~\cite{xu2023frozenrecon}} &\checkmark &
        & 0.1243 & 0.0762 & 0.0554 & 0.0231 & 0.6189  & 0.6885  & 0.2855  & 0.2059  & 0.1505 & 0.0631  & 0.6547  & 0.7577 & \footnotesize{\textless}0.1 \\
      \textbf{Dust3R$^\dagger$~\cite{wang2024dust3r}} &\checkmark &
        & \bf 0.0286 & \bf 0.0123 & 0.0280 & 0.0091 & \bf 0.6681   & \bf 0.7683  & \bf 0.0544  & \bf 0.0251  & 0.0315  & \bf 0.0103  & \bf 0.8024  & \bf 0.9529  & 0.78 \\
      \textbf{Ours} & & \checkmark
        & 0.0342 & 0.0148 & \bf 0.0241 & \bf 0.0085 & 0.6635  & 0.7625  & 0.0691  & 0.0315  & \bf 0.0291  & 0.0110  & 0.7775  & 0.9371  & \bf 65.49 \\
      \midrule
      \textbf{Dust3R~\cite{wang2024dust3r} (FV)} &\checkmark &
        & \bf 0.0188 & \bf 0.0087 & \bf 0.0234 & \bf 0.0096 & \bf 0.7851   & 0.8985  & \bf 0.0392   &\bf  0.0167 & \bf 0.0342 &\bf  0.0121 & \bf 0.8765   & \bf 0.9757   & 0.48 \\ 
      \textbf{Dust3R$^\dagger$~\cite{wang2024dust3r} (FV)} &\checkmark &
        & 0.0279 & 0.0133 & 0.0276 & 0.0108 & 0.7630   & 0.8841  & 0.0591  & 0.0266  & 0.0409  & 0.0136  & 0.8305  & 0.9556  & 1.42 \\ 
      \textbf{Ours$^\star$ (FV)} & &
        & 0.0233 &  0.0108 & 0.0246 & 0.0104 & 0.7791  & \bf 0.9003  & 0.0587  & 0.0239 & 0.0390  & 0.0132  & 0.8384  & 0.9616  & 5.83\\ 
      \textbf{Ours (FV)} & & \checkmark
        & 0.0239 & 0.0111 & 0.0247 & 0.0103 & 0.7768  & 0.8985  & 0.0611  & 0.0254  & 0.0392  & 0.0135  & 0.8330  & 0.9593  & \bf 72.04\\
      \bottomrule
    \end{tabularx}%
    \vspace{-5pt}

    \caption{\textbf{Quantitative results on 7Scenes~\cite{shotton2013scene} and NRGBD~\cite{Azinovic_2022_CVPR} datasets.} DUSt3R$^\dagger$ indicates using DUSt3R's final weights on $224\times 224$ images, same as our input resolution, to fit within 24GB GPU memory. For few-view (FV) reconstruction, we use the 8-frame pairs~\cite{duzceker2021deepvideomvs} as in SimpleRecon~\cite{sayed2022simplerecon}. Note that evaluating DUSt3R at the original resolution may benefit from increased visual overlapping.} 
    \vspace{-10pt}
    \label{tab:quantitative}
\end{table*}

%% file: Tables/quant_dtu.tex
\begin{table}[t]
  \centering
  \footnotesize
  \setlength{\tabcolsep}{0.3em}
    \begin{tabularx}{\columnwidth}{r c c >{\centering\arraybackslash}X >{\centering\arraybackslash}X >{\centering\arraybackslash}X >{\centering\arraybackslash}X >{\centering\arraybackslash}X >{\centering\arraybackslash}X}
      \toprule

         \multirow{2}{*}{Method} & \multicolumn{1}{c}{\multirow{2}{*}{Opt.}} & \multicolumn{1}{c}{\multirow{2}{*}{Onl.}} & \multicolumn{2}{c}{\tt{Acc}$\downarrow$} & \multicolumn{2}{c}{\tt{Comp}$\downarrow$} & \multicolumn{2}{c}{\tt{NC}$\uparrow$} \\ 
      \cmidrule(lr){4-5} \cmidrule(lr){6-7} \cmidrule(lr){8-9}
         & & & \tt{Mean} & \tt{Med.} & \tt{Mean} & \tt{Med.} & \tt{Mean} & \tt{Med.} \\
      \midrule
      
      \textbf{Dust3R~\cite{wang2024dust3r}} &\checkmark &
        & \bf 2.114 &  \bf 1.159 & \bf 2.033 & \bf 0.914 & \bf 0.749 & \bf 0.849\\
      \textbf{Dust3R$^\dagger$~\cite{wang2024dust3r}} &\checkmark &
        & 2.296 &  1.297 & 2.158 & 1.002 & 0.747 & 0.848\\
      \textbf{Ours$^\star$} & &
        & 2.902 & 1.273 & 2.120 & 0.937 & 0.732 & 0.836\\
        \textbf{Ours} & & \checkmark
        & 4.785 & 2.268 & 2.743 & 1.295 & 0.721 & 0.823\\
      \midrule
      \textbf{Dust3R~\cite{wang2024dust3r} (FV)} &\checkmark &
        & \bf 2.128 & \bf 1.241 & \bf 2.464 &  \bf 1.228 & \bf 0.797 & \bf 0.889 \\
      \textbf{Dust3R$^\dagger$~\cite{wang2024dust3r} (FV)} &\checkmark &
        & 2.511 & 1.484 & 2.661 &1.230 & 0.788 &  0.883 \\
      \textbf{Ours$^\star$ (FV)} & &
        & 3.055 & 1.600 & 2.878 & 1.345 & 0.781 & 0.878\\
      \textbf{Ours (FV)} & & \checkmark
        & 3.375 & 1.782 & 2.870 & 1.338 & 0.777 & 0.875\\
      \bottomrule
    \end{tabularx}
    \vspace{-5pt}

    \caption{\textbf{Quantitative results on DTU~\cite{aanaes2016large} dataset.} For few-view (FV) reconstruction, we use pairs provided in MVSNet~\cite{yao2018mvsnet}.} 
    \vspace{-5pt}
    \label{tab:quant_dtu}
\end{table}

%% file: Tables/quant_mem.tex
\begin{table}[t]
  \centering
  \footnotesize
  \setlength{\tabcolsep}{0.3em}
    \begin{tabularx}{\columnwidth}{r > {\centering\arraybackslash}X >{\centering\arraybackslash}X >{\centering\arraybackslash}X >{\centering\arraybackslash}X >{\centering\arraybackslash}X >{\centering\arraybackslash}X}
      \toprule

         \multirow{2}{*}{Method} & \multicolumn{2}{c}{\tt{Acc}$\downarrow$} & \multicolumn{2}{c}{\tt{Comp}$\downarrow$} & \multicolumn{2}{c}{\tt{NC}$\uparrow$} \\ 
      \cmidrule(lr){2-3} \cmidrule(lr){4-5} \cmidrule(lr){6-7}
         & \tt{Mean} & \tt{Med.} & \tt{Mean} & \tt{Med.} & \tt{Mean} & \tt{Med.} \\
      \midrule
      \textbf{w/o lm} & 0.2554 & 0.1419 & 0.1470 & 0.0872 & 0.5964  & 0.6523\\
      \textbf{w/o clip} & 0.0349 & 0.0161 & 0.0249 & 0.0090 & 0.6627  & 0.7614\\
      \textbf{Full} & \bf 0.0342 & \bf 0.0148 & \bf 0.0241 & \bf 0.0085 & \bf 0.6635  & \bf 0.7625\\
      \bottomrule
    \end{tabularx}%
    \vspace{-5pt}

    \caption{\textbf{Ablation studies on spatial memory.} w/o lm: use working memory only. w/o clipping the attention weight.} 
    \vspace{-5pt}
    \label{tab:ab_memory}
\end{table}

%% file: Figures/map.tex
\begin{figure*}[t]
    \centering
    \footnotesize
    \setlength{\tabcolsep}{1.2pt}
    \newcommand{\imgwidth}{0.16\textwidth} 

    \begin{tabular}{cccccc}
        \texttt{MA(1/3)} & \texttt{MA(2/3)} & \texttt{MA(3/3)} & \texttt{Kit.(1/3)} & \texttt{Kit.(2/3)} & \texttt{Kit.(3/3)} \\

        \includegraphics[width=\imgwidth]{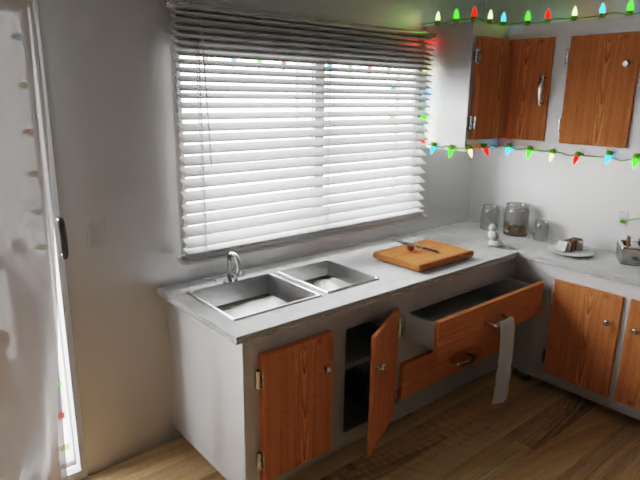} &
        \includegraphics[width=\imgwidth]{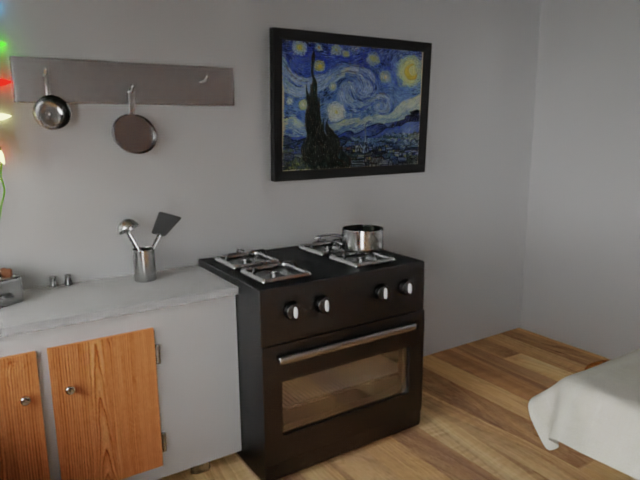} &
        \includegraphics[width=\imgwidth]{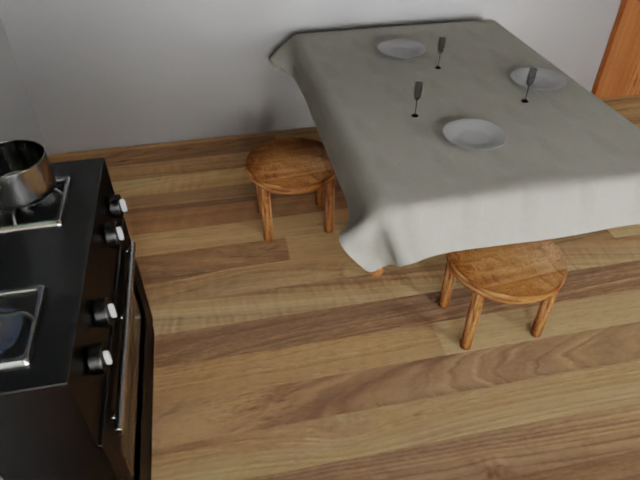} &
        \includegraphics[width=\imgwidth]{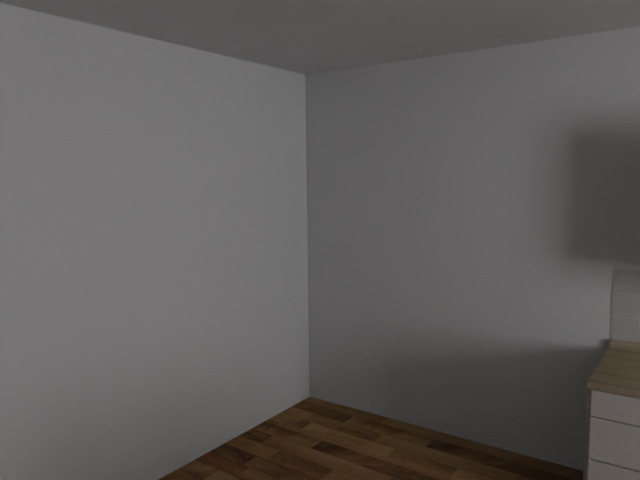} &
        \includegraphics[width=\imgwidth]{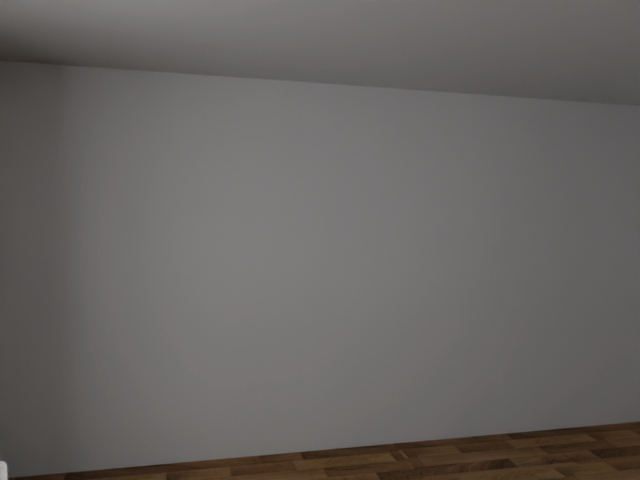} &
        \includegraphics[width=\imgwidth]{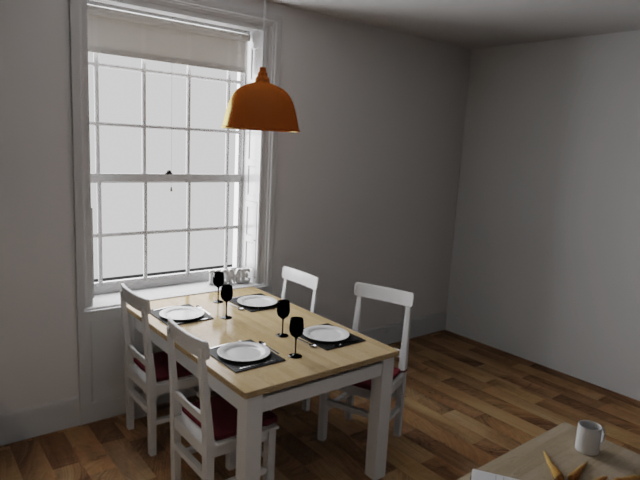} \\
        
        \includegraphics[width=\imgwidth]{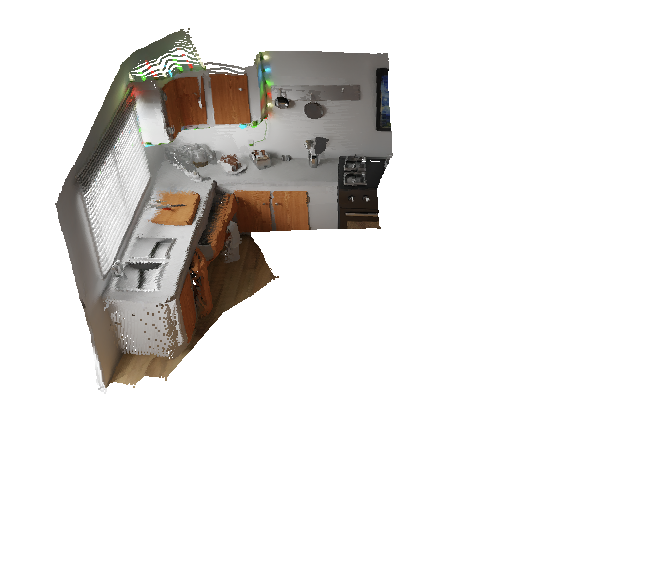} &
        \includegraphics[width=\imgwidth]{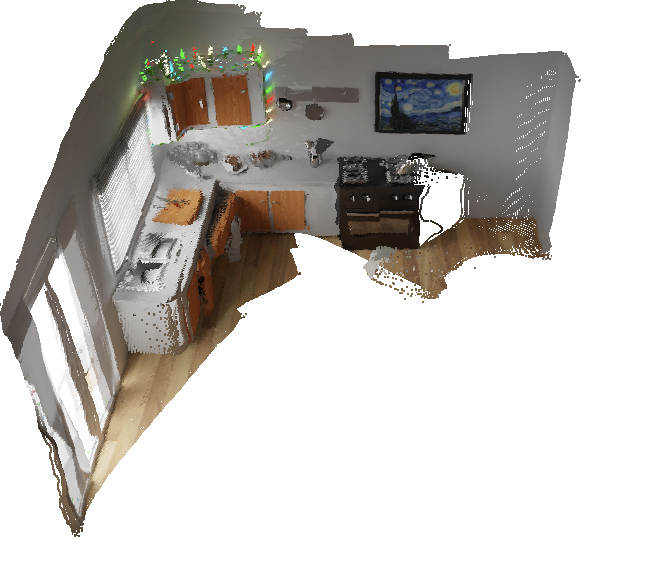} &
        \includegraphics[width=\imgwidth]{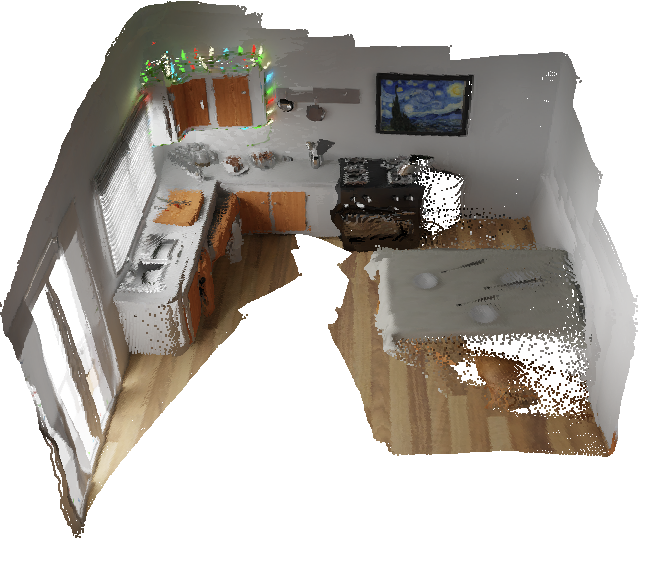} &
        \includegraphics[width=\imgwidth]{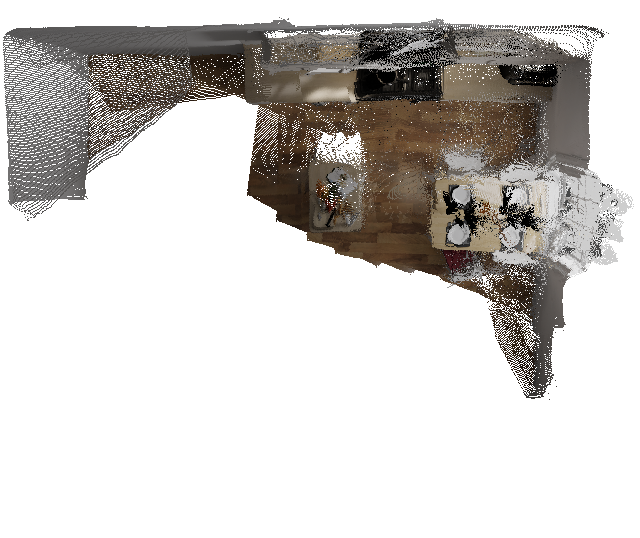} &
        \includegraphics[width=\imgwidth]{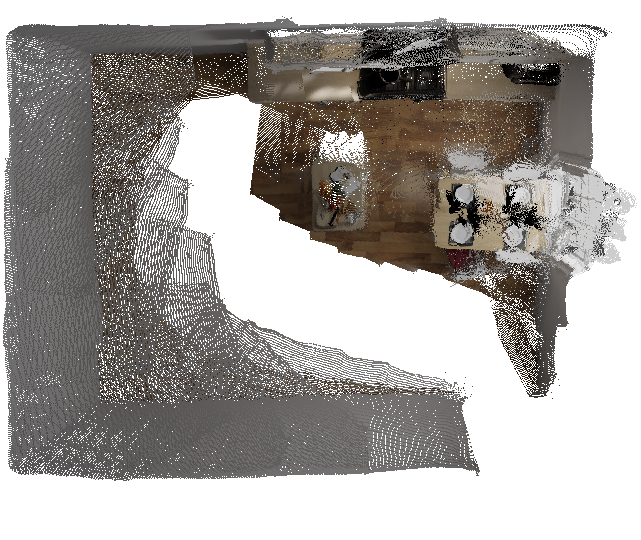} &
        \includegraphics[width=\imgwidth]{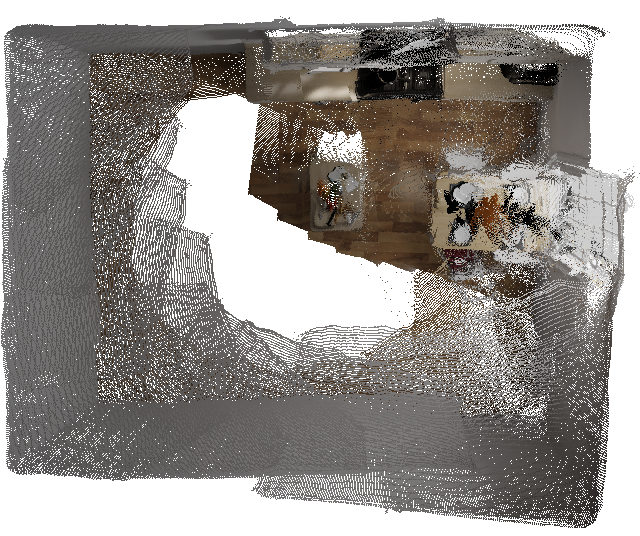} \\
    \end{tabular}
    \vspace{-5pt}
    \caption{\textbf{Online reconstruction.} We visualize the process of our online reconstruction in two indoor scenes. In both cases, our model shows its understanding of the regularity of the indoor scene, i.e., the Manhattan World Assumption. Our model can infer the geometry of the textureless wall based on those learned regularity. However, during loop closing, our model may not fill the geometry accurately due to the accumulated errors and outliers (noisy points around the window in the second scene.) }
    \label{fig:map}
    \vspace{-10pt}
\end{figure*}


%% file: Figures/memory_size.tex
\begin{figure}[t]
    \centering
    \includegraphics[width=0.99\columnwidth]{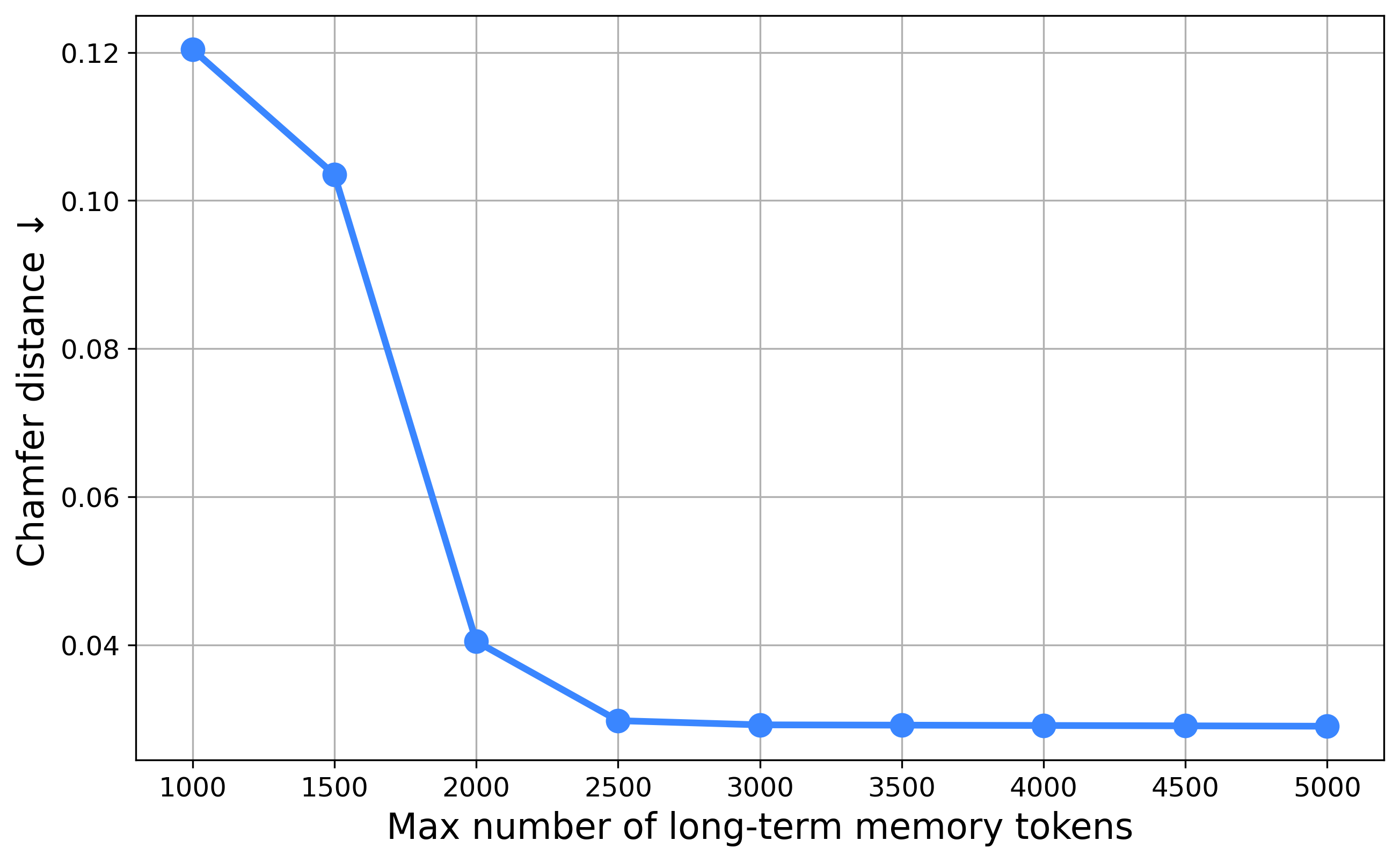}
    \vspace{-5pt}
    \caption{\textbf{Ablation study on memory size.} We plot Chamfer distance against the max number of tokens in long-term memory.}
    \label{fig:ab_memory}
    \vspace{-10pt}
\end{figure}

%% file: Figures/affinity.tex
\begin{figure}[t]
    \centering
    \setlength{\tabcolsep}{1pt}
    \newcommand{\imgwidth}{0.19\columnwidth}
    \begin{tabular}{ccccc}
        \includegraphics[width=\imgwidth]{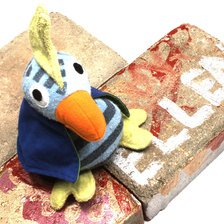} &
        \includegraphics[width=\imgwidth]{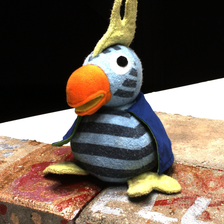} &
        \includegraphics[width=\imgwidth]{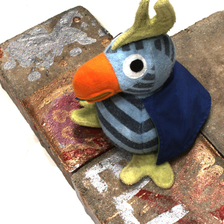} &
        \includegraphics[width=\imgwidth]{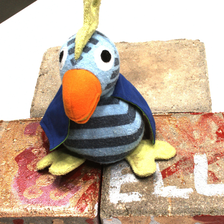} &
        \includegraphics[width=\imgwidth]{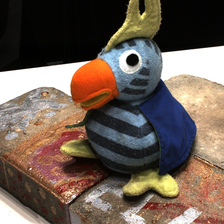} \\
        
        \includegraphics[width=\imgwidth]{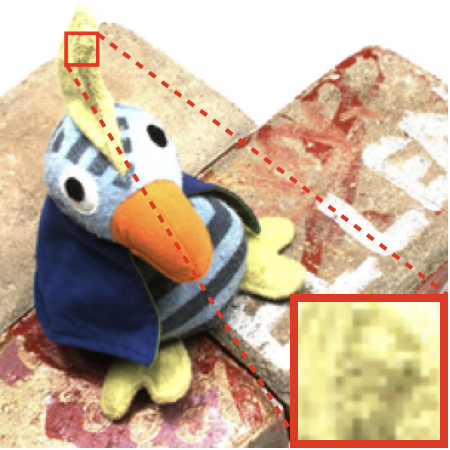} &
        \includegraphics[width=\imgwidth]{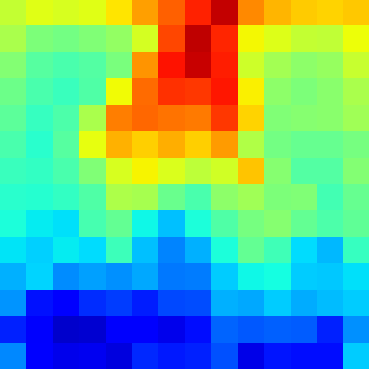} &
        \includegraphics[width=\imgwidth]{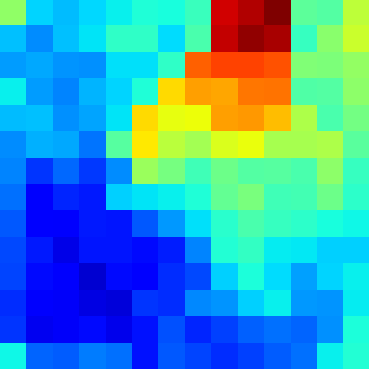} &
        \includegraphics[width=\imgwidth]{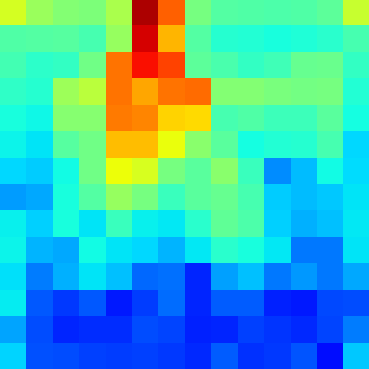} &
        \includegraphics[width=\imgwidth]{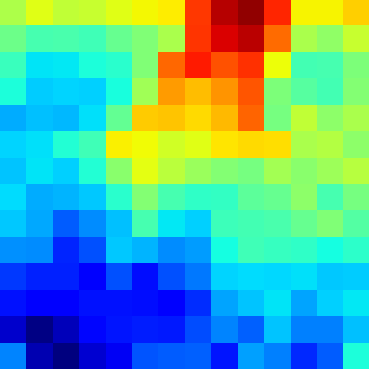} \\
        
        \includegraphics[width=\imgwidth]{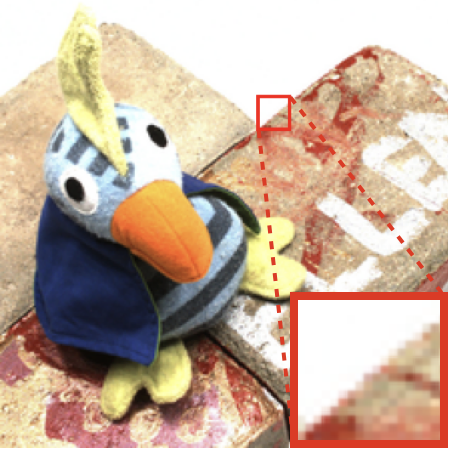} &
        \includegraphics[width=\imgwidth]{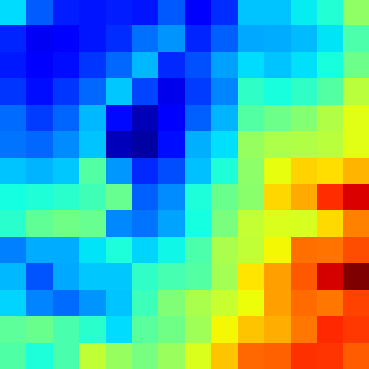} &
        \includegraphics[width=\imgwidth]{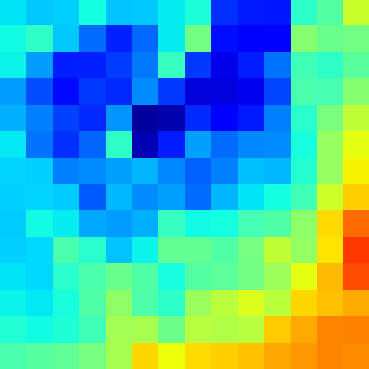} &
        \includegraphics[width=\imgwidth]{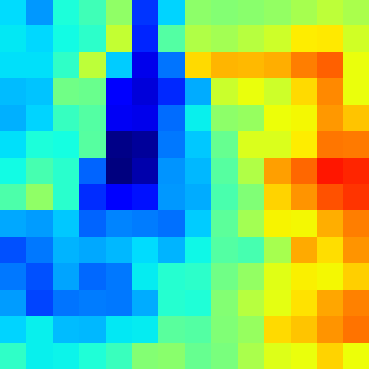} &
        \includegraphics[width=\imgwidth]{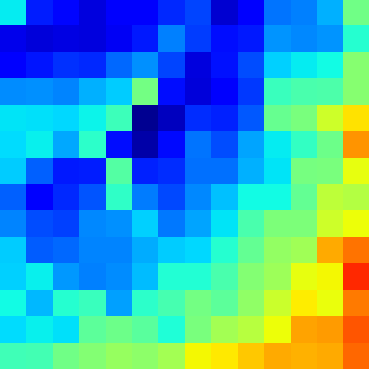} \\
        
        \includegraphics[width=\imgwidth]{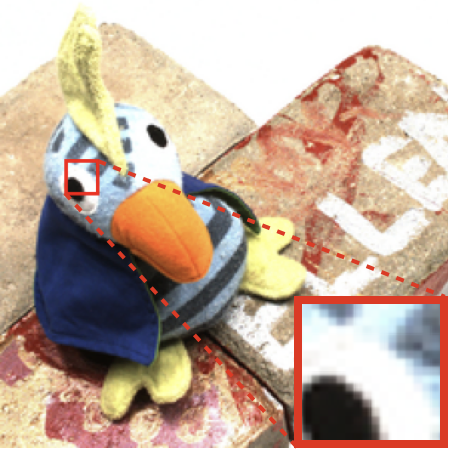} &
        \includegraphics[width=\imgwidth]{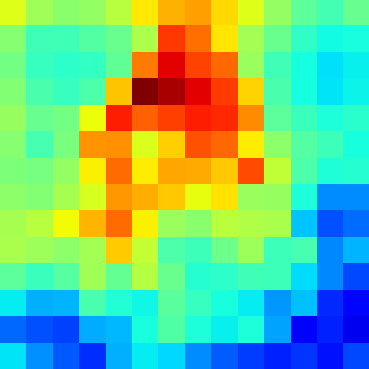} &
        \includegraphics[width=\imgwidth]{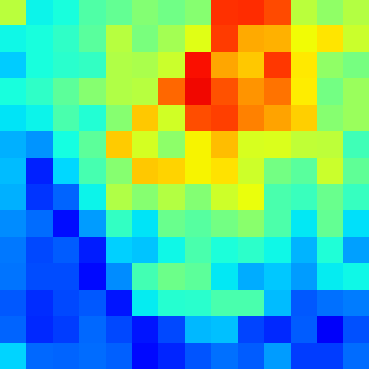} &
        \includegraphics[width=\imgwidth]{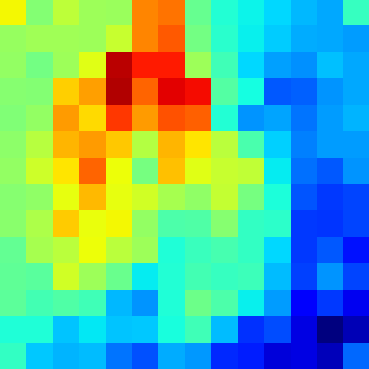} &
        \includegraphics[width=\imgwidth]{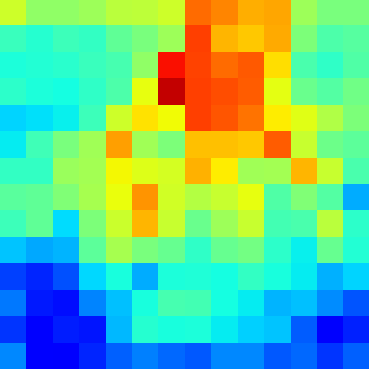} \\
        
        \includegraphics[width=\imgwidth]{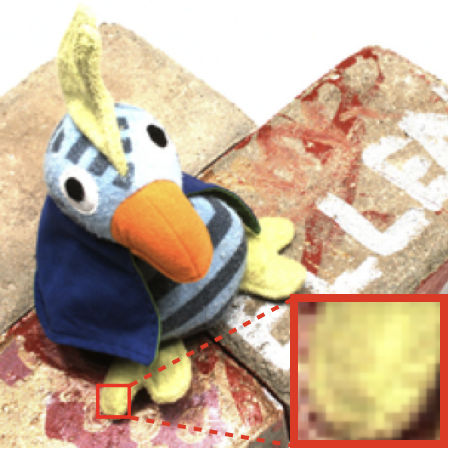} &
        \includegraphics[width=\imgwidth]{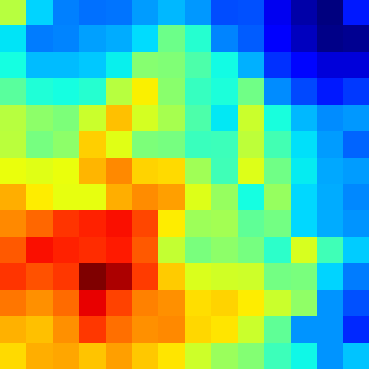} &
        \includegraphics[width=\imgwidth]{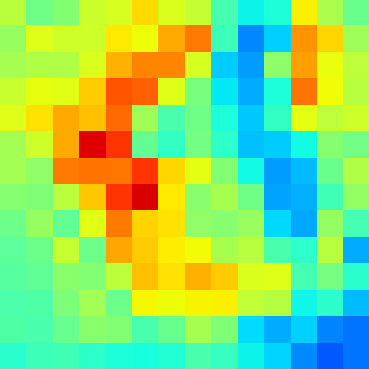} &
        \includegraphics[width=\imgwidth]{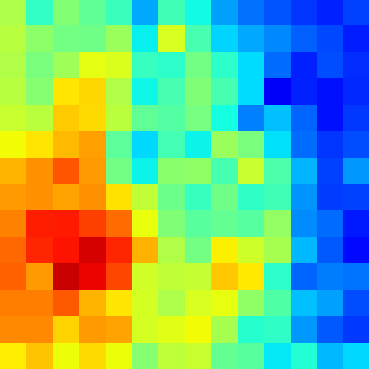} &
        \includegraphics[width=\imgwidth]{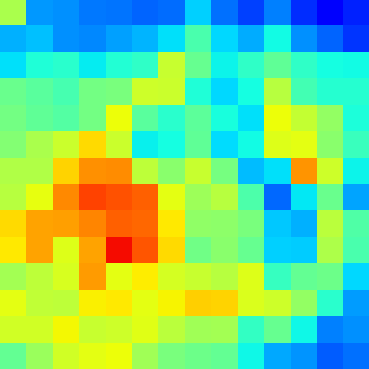} \\
    \end{tabular}
    \vspace{-5pt}
    \caption{\textbf{Visualization of the attention map.} We visualize the attention weight of selected patches with respect to all tokens in the memory. The results show robustness toward visually similar patches (e.g., right eye/feet). }
    \label{fig:attn}
    \vspace{-5pt}
\end{figure}

%% file: Figures/qual2.tex
\begin{figure*}[t]
    \centering
    \includegraphics[width=\textwidth]{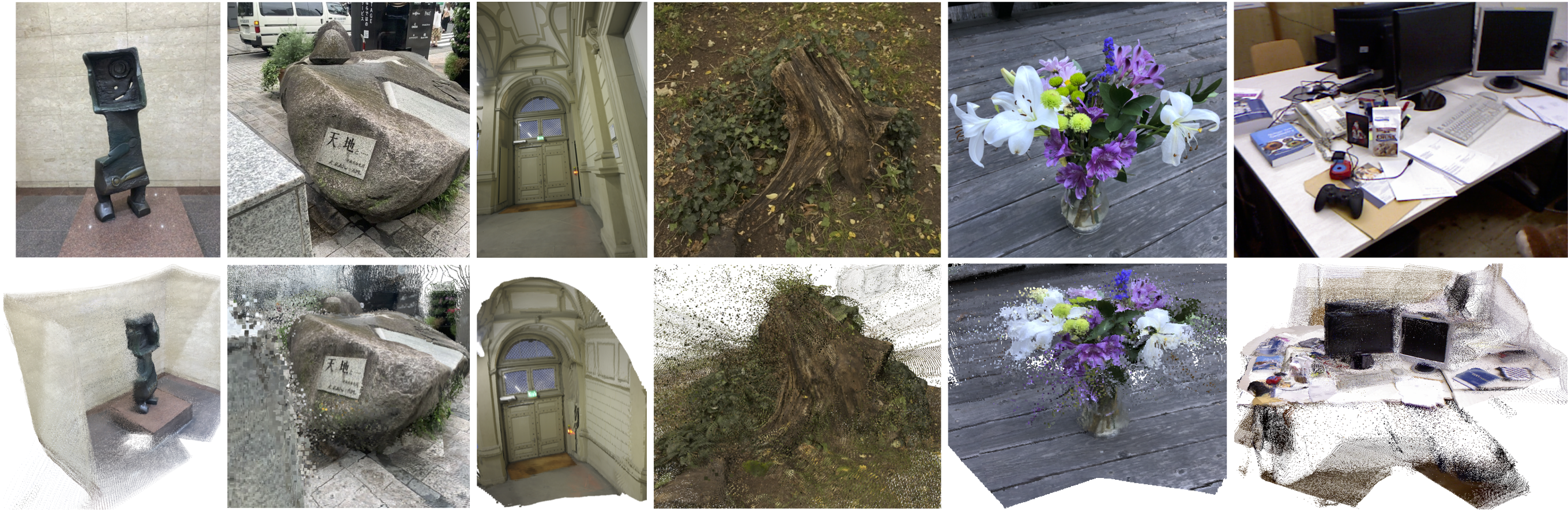}
    \vspace{-10pt}
    \caption{\textbf{Qualitative examples in various real-world datasets.} We visualize several reconstruction results of \papername{} on Map-free Reloc~\cite{arnold2022mapfree}, ETH3D~\cite{schops2017etb3d}, MipNeRF-360~\cite{barron2022mip}, NeRF~\cite{mildenhall2020nerf} and TUM-RGBD~\cite{sturm2012tumrgbd} datasets to demonstrate the generalization ability of our methods on different type of scenes, including indoor, outdoor, object-level, scene-level reconstruction. } 
    \label{fig:qual_gen}
    \vspace{-10pt}
\end{figure*}

%% file: sec/5_concl.tex
\section{Conclusion}

We have presented \papername{}, a model capable of achieving incremental reconstruction from RGB images without requiring prior knowledge of the camera parameters. By introducing the concept of spatial memory, which encodes previous states for next-frame prediction, \papername{} reconstructs scenes through a simple forward pass with a transformer-based architecture, eliminating the need for test-time optimization. This enables online reconstruction in real time. Trained on various large-scale datasets, \papername{} demonstrates competitive reconstruction quality and generalization ability across various scenarios. Future work includes extending our method to handle large-scale scenes, incorporating bundle adjustment techniques, and exploring self-supervised training on casual videos.

\noindent
\textbf{Acknowledgements.} The research presented here has been supported by a sponsored research award from Cisco Research and the UCL Centre for Doctoral Training in Foundational AI under UKRI grant number EP/S021566/1. This project made use of time on Tier 2 HPC facility JADE2, funded by EPSRC (EP/T022205/1). 

%% file: sec/X_suppl.tex
\clearpage
\maketitlesupplementary

\section{Additional details}

\noindent
\textbf{Training loss.} The confidence loss as in DUSt3R~\cite{wang2024dust3r} is:
\begin{equation}
    \gL_{\mathrm{conf}} = \sum_t \sum_{i\in\gV} C_t^i \gL_{\mathrm{reg}}(i) - \alpha \log C_t^i,
\end{equation}

\noindent
where $\gV$ is the set of all valid pixels. The confidence $C_t^i$ is an exponential function of the raw output of the network $\hat{C}_t^i$:
\begin{equation}
    C_t^i = 1 + \exp(\hat{C}_t^i)
    \label{eq:conf_exp}
\end{equation}

\noindent
In confidence loss $\gL_{\mathrm{conf}}$, $\alpha$ controls the total confidence score the model needs to distribute to the loss of each pixel. Since the regions with larger depths usually have a larger loss, the model assigns more confidence weight to regions with smaller depths. This loss shares a similar spirit to other depth representations, e.g., inverse depth, which gives more weight to pixels with smaller depth. However, instead of explicitly encoding the depth, this confidence loss let the model learn the weight function along the training. Our scale loss is defined as:
\begin{equation}
    \gL_{\mathrm{scale}} = \max(0, \bar{X}-\bar{X}_{\mathrm{gt}}),
\end{equation}

\noindent
where $\bar{X}$ and $\bar{X}_{\mathrm{gt}}$ are the average distance of all predicted and ground-truth points to the origin. The scale loss encourages the predicted scale to be smaller than the GT scale to prevent the model from learning trivial solutions.

To tune the hyper-parameter $\alpha$, we find that the best way is to ensure the overall training loss becomes smaller than $0$ after $30\%$ of epochs. A typical sign of choosing the $\alpha$ that is not big enough is the $\gL_{\mathrm{scale}}$ becomes quite large along the training. This indicates that some pixels with large depths make the model predict the trivial solutions. We find that $\alpha \geq 0.4$ achieves the best results in our case.

\noindent
\textbf{Curriculum training.} Given the minimal and maximum sampling interval $T_\mathrm{min}$ and $T_\mathrm{max}$ between adjacent frames, our curriculum sampling can be written as:

\begin{equation}
    T = T_\mathrm{min} + \eta_a (T_\mathrm{max} - T_\mathrm{min}),
\end{equation}

\noindent
where $\eta_a$ is the active ratio of the training ratio $\eta$:

\begin{equation}
    \eta_a =
\begin{cases} 
\min\left(1, 2\eta\right) & \text{if } \eta < 0.75 \\
\max\left(0.5, 4 - 4\eta\right) & \text{otherwise}
\end{cases}
\end{equation}

\section{Additional analysis}

\noindent
\textbf{Ablation study on view selection.} Since the confidence function in Eq.~\ref{eq:conf_exp} tends to over-weight patches with higher confidence, we instead use the sigmoid function for view selection of the offline reconstruction. The overall confidence function becomes:
\begin{equation}
 C =  \frac{C_1 -1}{C_1} + \frac{C_2 -1}{C_2}.
\end{equation}

\noindent
The difference in performance is illustrated in Tab.~\ref{tab:ab_conf}.

\input{Tables/quant_conf}

\input{Tables/quant_dust3r}

\input{Tables/quant_per_scene_7scenes}

\input{Figures/supp_gwr}
\input{Tables/quant_per_scene_nrgbd}

\noindent
\textbf{Ablation study on DUSt3R in \papername{}.} Since our model inherits from the network architecture of DUSt3R~\cite{wang2024dust3r}, we can directly compare the performance of the ViT encoder with two decoders in our model, denoted as DUSt3R$^\mathrm{ours}$, with the original DUSt3R. As shown in Tab.~\ref{tab:ab_dust_scene} and Tab.~\ref{tab:ab_dust_dtu}, even though we re-purpose the two decoders, DUSt3R$^\mathrm{ours}$ still shows on-par median accuracy and completion and consistent better normal consistency compared to DUSt3R$^\dagger$ on indoor scene reconstruction. This opens up the possibility of combining optimization-based techniques in DUSt3R with \papername{} within one set of model parameters. Additionally, the inferior results on DTU datasets might be due to 1) Our training set only consists of a small fraction of object-centric scenes. 2) DUSt3R uses an internal pair selection model, which can potentially boost the performance of the object-centric scenes. In contrast, we use a simple strategy of random sampling.

\noindent
\textbf{Per-scene performance.} We show a per-scene breakdown of quantitative results in Tab.~\ref{tab:per_scene_7scenes} and Tab.~\ref{tab:per_scene_NRGBD}. Our method achieves competitive per-scene results compared to DUSt3R. However, in some challenging scenes, our model might produce more outliers compared to DUSt3R, which leads to a higher accuracy score. Fig~\ref{fig:gwr_images} shows an example on the NRGBD dataset, where the scene contains a mirror. This leads our model to produce more outliers and eventually leads to twice higher accuracy compared to DUSt3R.

%% file: Tables/quant_conf.tex
\begin{table}[t]
  \centering
  \footnotesize
  \setlength{\tabcolsep}{0.3em}
    \begin{tabularx}{\columnwidth}{r > {\centering\arraybackslash}X >{\centering\arraybackslash}X >{\centering\arraybackslash}X >{\centering\arraybackslash}X >{\centering\arraybackslash}X >{\centering\arraybackslash}X}
      \toprule

         \multirow{2}{*}{Method} & \multicolumn{2}{c}{\tt{Acc}$\downarrow$} & \multicolumn{2}{c}{\tt{Comp}$\downarrow$} & \multicolumn{2}{c}{\tt{NC}$\uparrow$} \\ 
      \cmidrule(lr){2-3} \cmidrule(lr){4-5} \cmidrule(lr){6-7}
         & \tt{Mean} & \tt{Med.} & \tt{Mean} & \tt{Med.} & \tt{Mean} & \tt{Med.} \\
      \midrule
      \textbf{Ours$^\star$ (exp)} & 3.099 & 1.361 & 2.247 & 0.993 & 0.731 & 0.835\\
      \textbf{Ours$^\star$} & \bf 2.902 & \bf 1.273 & \bf 2.120 & \bf 0.937 & \bf 0.732 & \bf 0.836\\
      \bottomrule
    \end{tabularx}%
    \vspace{-5pt}

    \caption{\textbf{Ablation study on view selection.} Ours$^\star$ (exp): exponential confidence function for view selection as in DUSt3R~\cite{wang2024dust3r}. Ours$^\star$:sigmoid confidence function for view selection.} 
    \vspace{-5pt}
    \label{tab:ab_conf}
\end{table}

%% file: Tables/quant_dust3r.tex
\begin{table}[t]
  \centering
  \footnotesize
  \setlength{\tabcolsep}{0.3em}
    \begin{tabularx}{\columnwidth}{ll > {\centering\arraybackslash}X >{\centering\arraybackslash}X >{\centering\arraybackslash}X >{\centering\arraybackslash}X >{\centering\arraybackslash}X >{\centering\arraybackslash}X}
      \toprule

          \multirow{2}{*}{datasets} &\multirow{2}{*}{Method} & \multicolumn{2}{c}{\tt{Acc}$\downarrow$} & \multicolumn{2}{c}{\tt{Comp}$\downarrow$} & \multicolumn{2}{c}{\tt{NC}$\uparrow$} \\ 
      \cmidrule(lr){3-4} \cmidrule(lr){5-6} \cmidrule(lr){7-8}
         & & \tt{Mean} & \tt{Med.} & \tt{Mean} & \tt{Med.} & \tt{Mean} & \tt{Med.} \\
      \midrule
     \multirow{3}{*}{7scenes} & \textbf{Dust3R$^\dagger$} &0.0286 & 0.0123 & 0.0280 & 0.0091 & 0.6681 & 0.7683\\
     & \textbf{Dust3R$^\mathrm{ours}$} &\bf 0.0278 & \bf 0.0117 & 0.0247 & 0.0101 & \bf 0.6775   & \bf 0.7842\\
     
     &\textbf{Ours} & 0.0342 & 0.0148 & \bf 0.0241 & \bf 0.0085 & 0.6635  & 0.7625\\

      \midrule
     \multirow{3}{*}{\begin{tabular}{@{}c@{}}7scenes\\(FV)\end{tabular}} & \textbf{Dust3R$^\dagger$} &0.0279 & 0.0133 & 0.0276 & 0.0108 & 0.7630   & 0.8841\\
     
     & \textbf{Dust3R$^\mathrm{ours}$} &0.0242 & 0.0114 & 0.0249 & 0.0106 & \bf 0.7785   & \bf 0.9003\\

     & \textbf{Ours} & \bf 0.0239 & \bf 0.0111 & \bf 0.0247 & \bf 0.0103 & 0.7768  & 0.8985\\

     \midrule
      \multirow{3}{*}{NRGBD} & \textbf{Dust3R$^\dagger$} &\bf 0.0544  & 0.0251  & 0.0315  & \bf 0.0103  & 0.8024  & 0.9529\\
      & \textbf{Dust3R$^\mathrm{ours}$} & 0.0644 & \bf 0.0246 & 0.0396 & 0.0110 & \bf 0.8041   & \bf \bf 0.9623\\

      & \textbf{Ours} & 0.0691  & 0.0315  & \bf 0.0291  & 0.0110  & 0.7775  & 0.9371\\ 

     \midrule
     \multirow{3}{*}{\begin{tabular}{@{}c@{}}NRGBD\\(FV)\end{tabular}} & \textbf{Dust3R$^\dagger$} &\bf 0.0591  & 0.0266  & 0.0409  & 0.0136  & 0.8305  & 0.9556\\
     & \textbf{Dust3R$^\mathrm{ours}$} &0.0606 & \bf 0.0252 & 0.0407 & 0.0143 & \bf 0.8439   & \bf 0.9630\\
     & \textbf{Ours} & 0.0611  & 0.0254  & \bf 0.0392  & \bf 0.0135  & 0.8330  & 0.9593 \\

      \bottomrule
    \end{tabularx}%
    \vspace{-5pt}

    \caption{\textbf{Ablation study on Dust3R$^\mathrm{ours}$ on indoor scene.}} 
    \vspace{-5pt}
    \label{tab:ab_dust_scene}
\end{table}

\begin{table}[t]
  \centering
  \footnotesize
  \setlength{\tabcolsep}{0.3em}
    \begin{tabularx}{\columnwidth}{ll > {\centering\arraybackslash}X >{\centering\arraybackslash}X >{\centering\arraybackslash}X >{\centering\arraybackslash}X >{\centering\arraybackslash}X >{\centering\arraybackslash}X}
      \toprule

          \multirow{2}{*}{datasets} &\multirow{2}{*}{Method} & \multicolumn{2}{c}{\tt{Acc}$\downarrow$} & \multicolumn{2}{c}{\tt{Comp}$\downarrow$} & \multicolumn{2}{c}{\tt{NC}$\uparrow$} \\ 
      \cmidrule(lr){3-4} \cmidrule(lr){5-6} \cmidrule(lr){7-8}
         & & \tt{Mean} & \tt{Med.} & \tt{Mean} & \tt{Med.} & \tt{Mean} & \tt{Med.} \\

      \midrule
      \multirow{3}{*}{DTU} &\textbf{Dust3R$^\dagger$}
      & \bf 2.296 &  \bf 1.297 & 2.158 & \bf 1.002 & \bf 0.747 & \bf 0.848\\

      &\textbf{Dust3R$^\mathrm{ours}$}
      & 3.386 &  1.469 & 2.228 & 1.017 & 0.734 & 0.837\\
      & \textbf{Ours$^\star$} & 2.902 & 1.273 & \bf 2.120 & 0.937 & 0.732 & 0.836\\

     \midrule
      \multirow{3}{*}{\begin{tabular}{@{}c@{}}DTU\\(FV)\end{tabular}} &\textbf{Dust3R$^\dagger$}
      & \bf 2.511 & \bf 1.484 & \bf 2.661 & \bf 1.230 & \bf 0.788 & \bf 0.883\\
      &\textbf{Dust3R$^\mathrm{ours}$}
      & 3.875 &  1.869 & 2.916 & 1.438 & 0.777 & 0.874\\
      & \textbf{Ours$^\star$ (FV)} & 3.055 & 1.600 & 2.878 & 1.345 & 0.781 & 0.878\\

      \bottomrule
    \end{tabularx}%
    \vspace{-5pt}

    \caption{\textbf{Ablation study on Dust3R$^\mathrm{ours}$ on DTU.}} 
    \vspace{-8pt}
    \label{tab:ab_dust_dtu}
\end{table}

%% file: Tables/quant_per_scene_7scenes.tex
\begin{table}[t]
  \centering
  \footnotesize
  \setlength{\tabcolsep}{0.3em}
    \begin{tabularx}{\columnwidth}{ll >{\centering\arraybackslash}X >{\centering\arraybackslash}X >{\centering\arraybackslash}X >{\centering\arraybackslash}X >{\centering\arraybackslash}X >{\centering\arraybackslash}X}
      \toprule
          \multirow{2}{*}{Scene} & \multirow{2}{*}{Method} & \multicolumn{2}{c}{\tt{Acc}$\downarrow$} & \multicolumn{2}{c}{\tt{Comp}$\downarrow$} & \multicolumn{2}{c}{\tt{NC}$\uparrow$} \\ 
      \cmidrule(lr){3-4} \cmidrule(lr){5-6} \cmidrule(lr){7-8}
         & & \tt{Mean} & \tt{Med.} & \tt{Mean} & \tt{Med.} & \tt{Mean} & \tt{Med.} \\
      \midrule
      \multirow{2}{*}{chess03} & \textbf{Dust3R$^\dagger$} & 0.0270 & 0.0093 & \bf 0.0180 & 0.0055 & 0.6351 & 0.7144\\
                                    & \textbf{Ours}             & \bf 0.0237 & \bf 0.0072 & 0.0193 & \bf 0.0052 & \bf 0.6505 & \bf 0.7389\\
      \midrule
      \multirow{2}{*}{chess05} & \textbf{Dust3R$^\dagger$} & 0.0335 & 0.0141 & 0.0178 & 0.0080 & 0.6352 & 0.7156\\
                                    & \textbf{Ours}             & \bf 0.0229 & \bf 0.0073 & \bf 0.0142 & \bf 0.0058 & \bf 0.6413 & \bf 0.7249\\
      \midrule
      \multirow{2}{*}{pumpkin01} & \textbf{Dust3R$^\dagger$} & 0.0337 & 0.0133 & 0.0292 & 0.0123 & \bf 0.7302 & \bf 0.8509\\
                                      & \textbf{Ours}             & \bf 0.0271 & \bf 0.0131 & \bf 0.0205 & \bf 0.0087 & 0.7068 & 0.8258\\
      \midrule
      \multirow{2}{*}{pumpkin07} & \textbf{Dust3R$^\dagger$} & 0.0193 & \bf 0.0055 & \bf 0.0132 & \bf 0.0052 & 0.6793 & 0.7860\\
                                      & \textbf{Ours}             & \bf 0.0178 & 0.0062 & 0.0164 & 0.0060 & \bf 0.6832 & \bf 0.7929\\
      \midrule
      \multirow{2}{*}{stairs01} & \textbf{Dust3R$^\dagger$} & \bf 0.0636 & \bf 0.0357 & 0.1023 & 0.0193 & 0.6475 & 0.7476\\
                                    & \textbf{Ours}             & 0.0739 & 0.0421 & \bf 0.0672 & \bf 0.0151 & \bf 0.6507 & \bf 0.7496\\
      \midrule
      \multirow{2}{*}{stairs04} & \textbf{Dust3R$^\dagger$} & 0.0475 & 0.0212 & 0.0900 & 0.0174 & 0.6446 & 0.7335\\
                                    & \textbf{Ours}             & \bf 0.0390 & \bf 0.0160 & \bf 0.0357 & \bf 0.0069 & \bf 0.6588 & \bf 0.7589\\
      \midrule
      \multirow{2}{*}{fire03}   & \textbf{Dust3R$^\dagger$} & 0.0112 & 0.0044 & 0.0096 & 0.0042 & \bf 0.6539 & \bf 0.7474\\
                                     & \textbf{Ours}             & \bf 0.0089 & \bf 0.0042 & \bf 0.0086 & \bf 0.0039 & 0.6523 & 0.7454\\
      \midrule
      \multirow{2}{*}{fire04}   & \textbf{Dust3R$^\dagger$} & 0.0104 & 0.0037 & 0.0111 & 0.0037 & 0.6515 & 0.7408\\
                                     & \textbf{Ours}             & \bf 0.0086 & \bf 0.0034 & \bf 0.0098 & \bf 0.0036 & \bf 0.6556 & \bf 0.7472\\
      \midrule
      \multirow{2}{*}{office02} & \textbf{Dust3R$^\dagger$} & 0.0462 & \bf 0.0179 & 0.0381 & 0.0145 & 0.6819 & 0.7957\\
                                     & \textbf{Ours}             & \bf 0.0403 & 0.0187 & \bf 0.0223 & \bf 0.0124 & \bf 0.6843 & \bf 0.7969\\
      \midrule
      \multirow{2}{*}{office06} & \textbf{Dust3R$^\dagger$} & \bf 0.0257 & \bf 0.0152 & \bf 0.0218 & \bf 0.0094 & \bf 0.7195 & \bf 0.8477\\
                                     & \textbf{Ours}             & 0.0879 & 0.0414 & 0.0420 & 0.0154 & 0.6731 & 0.7823\\
      \midrule
      \multirow{2}{*}{office07} & \textbf{Dust3R$^\dagger$} & 0.0270 & 0.0132 & \bf 0.0224 & 0.0126 & \bf 0.6864 & \bf 0.7944\\
                                     & \textbf{Ours}             & \bf 0.0269 & \bf 0.0127 & 0.0232 & \bf 0.0101 & 0.6740 & 0.7772\\
      \midrule
      \multirow{2}{*}{office09} & \textbf{Dust3R$^\dagger$} & \bf 0.0351 & \bf 0.0165 & \bf 0.0281 & \bf 0.0102 & \bf 0.6777 & \bf 0.7854\\
                                     & \textbf{Ours}             & 0.0791 & 0.0279 & 0.0541 & 0.0192 & 0.6579 & 0.7560\\
      \midrule
      \multirow{2}{*}{redkit03} & \textbf{Dust3R$^\dagger$} & \bf 0.0250 & \bf 0.0112 & 0.0183 & 0.0087 & \bf 0.6983 & \bf 0.8186\\
                                         & \textbf{Ours}             & 0.0367 & 0.0203 & \bf 0.0158 & \bf 0.0075 & 0.6765 & 0.7849\\
      \midrule
      \multirow{2}{*}{redkit04} & \textbf{Dust3R$^\dagger$} & \bf 0.0184 & \bf 0.0069 & 0.0235 & \bf 0.0059 & \bf 0.6570 & \bf 0.7509\\
                                         & \textbf{Ours}             & 0.0242 & 0.0083 & \bf 0.0179 & 0.0061 & 0.6532 & 0.7467\\
      \midrule
      \multirow{2}{*}{redkit06} & \textbf{Dust3R$^\dagger$} &\bf  0.0240 & 0.0127 & \bf 0.0170 &\bf  0.0075 & \bf 0.6533 & \bf 0.7442\\
                                         & \textbf{Ours}             & 0.0285 & \bf 0.0120 & 0.0214 & 0.0087 & 0.6404 & 0.7229\\
      \midrule
      \multirow{2}{*}{redkit12} & \textbf{Dust3R$^\dagger$} & \bf 0.0191 & \bf 0.0068 & \bf 0.0161 & \bf 0.0074 & \bf 0.6423 & \bf 0.7271\\
                                         & \textbf{Ours}             & 0.0257 & 0.0091 & 0.0178 & 0.0076 & 0.6364 & 0.7211\\
      \midrule
      \multirow{2}{*}{redkit14} & \textbf{Dust3R$^\dagger$} & 0.0216 & 0.0087 & 0.0197 & 0.0080 & 0.6332 & 0.7106\\
                                         & \textbf{Ours}             & \bf 0.0187 & \bf 0.0086 & \bf 0.0171 & \bf 0.0070 & \bf 0.6427 & \bf 0.7264\\
      \midrule
      \multirow{2}{*}{heads01} & \textbf{Dust3R$^\dagger$} & \bf 0.0256 & \bf 0.0056 & \bf 0.0082 & \bf 0.0037 & 0.6983 & 0.8180\\
                                    & \textbf{Ours}             & 0.0267 & 0.0082 & 0.0098 & 0.0043 & \bf 0.7056 & \bf 0.8288\\
      \midrule
      \multirow{2}{*}{Avg.}      & \textbf{Dust3R$^\dagger$} & \bf 0.0286 & \bf 0.0123 & 0.0280 & 0.0091 & \bf 0.6681 & \bf 0.7683\\
                                    & \textbf{Ours}             & 0.0342 & 0.0148 & \bf 0.0241 & \bf 0.0085 & 0.6635 & 0.7625\\
      \bottomrule
    \end{tabularx}%
    \vspace{-5pt}
    \caption{\textbf{Per-scene results on 7scenes dataset.}} 
    \vspace{-8pt}
    \label{tab:per_scene_7scenes}
\end{table}

%% file: Figures/supp_gwr.tex
\begin{figure}[t]
    \centering
    \footnotesize
    \setlength{\tabcolsep}{1pt}
    \newcommand{\imgwidth}{0.32\columnwidth}
    \begin{tabular}{ccc}
        
        DUSt3R$^\dagger$ & DUSt3R$^\mathrm{ours}$ & Ours\\
        Acc: 0.1096 & Acc: 0.1986 & Acc:  0.2074\\
        
        \includegraphics[width=\imgwidth]{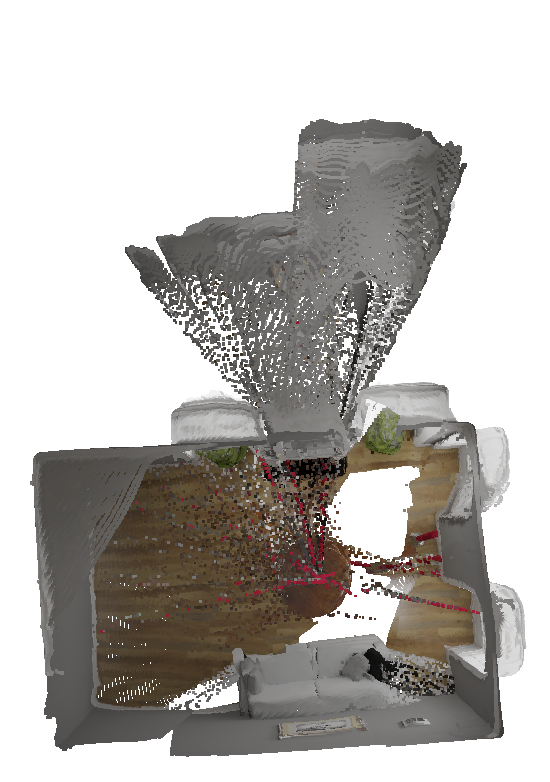} &
        \includegraphics[width=\imgwidth]{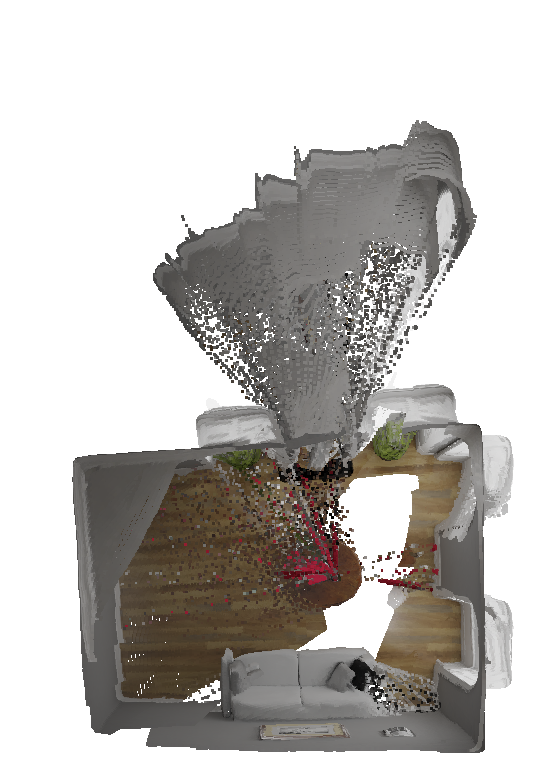} &
        \includegraphics[width=\imgwidth]{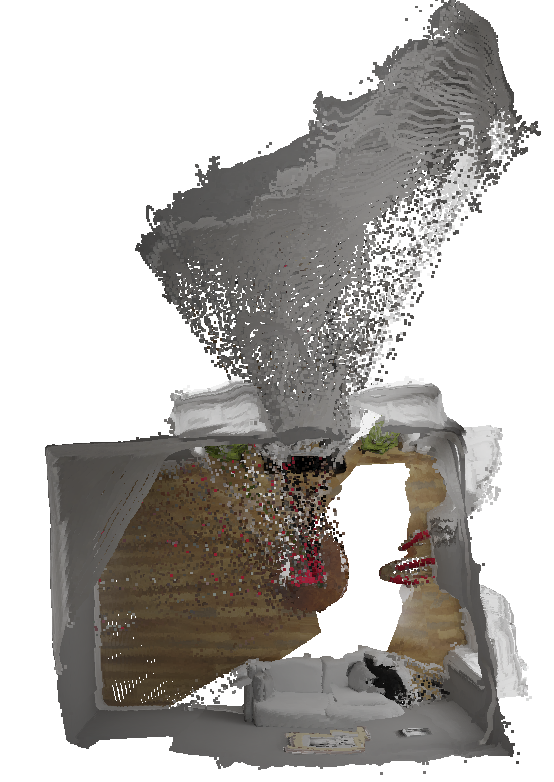} \\
        
        \includegraphics[width=\imgwidth]{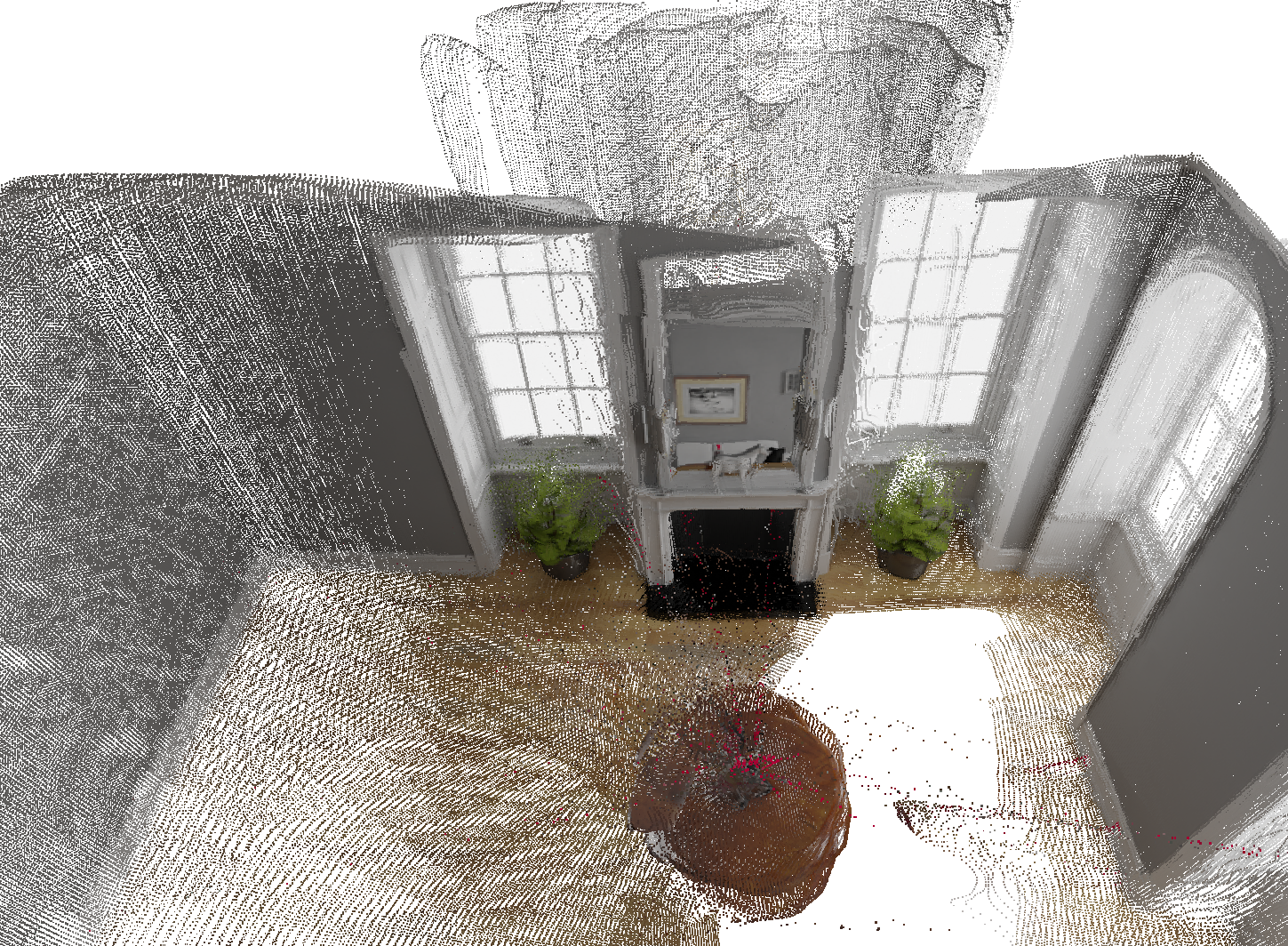} &
        \includegraphics[width=\imgwidth]{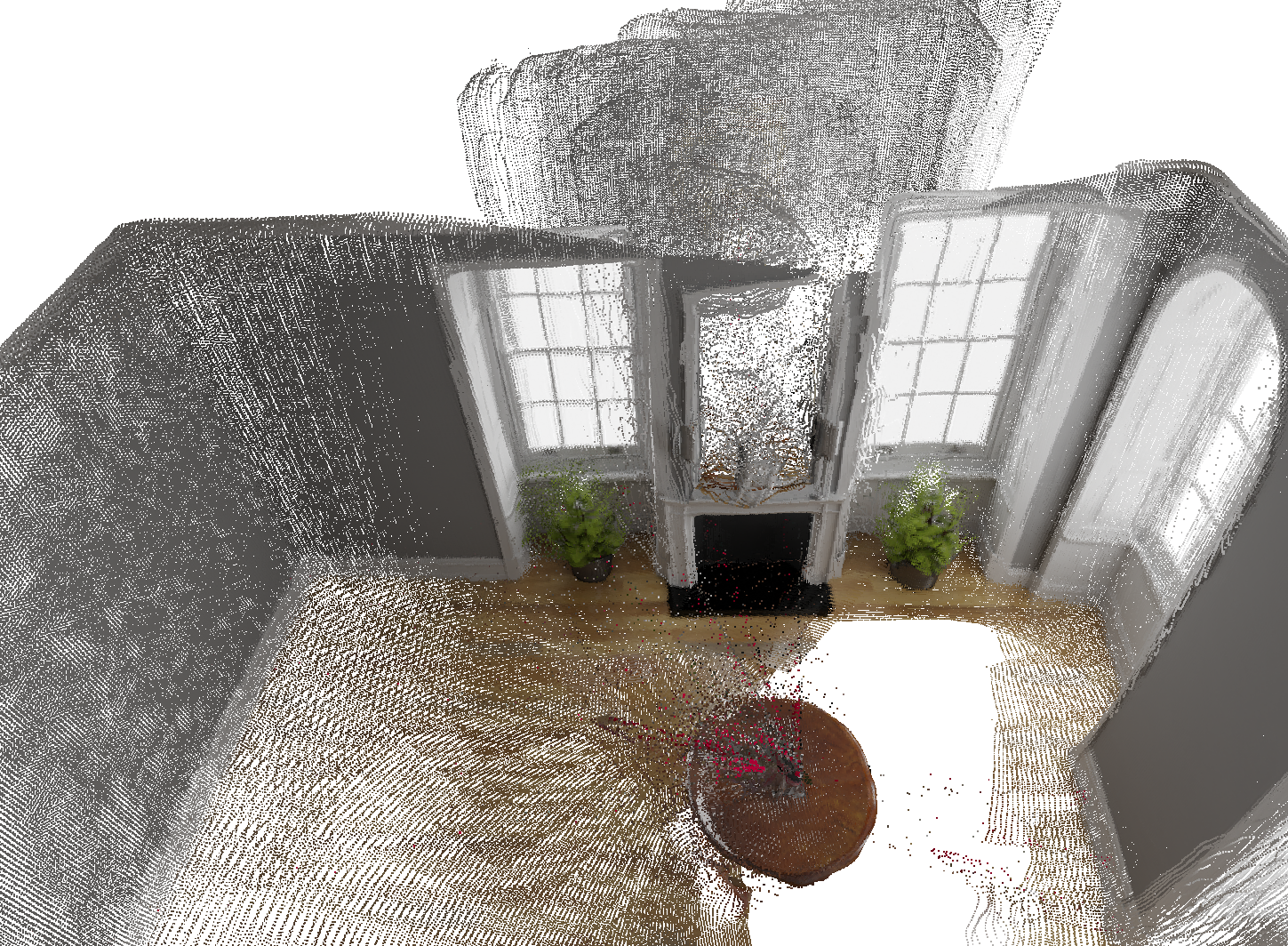} &
        \includegraphics[width=\imgwidth]{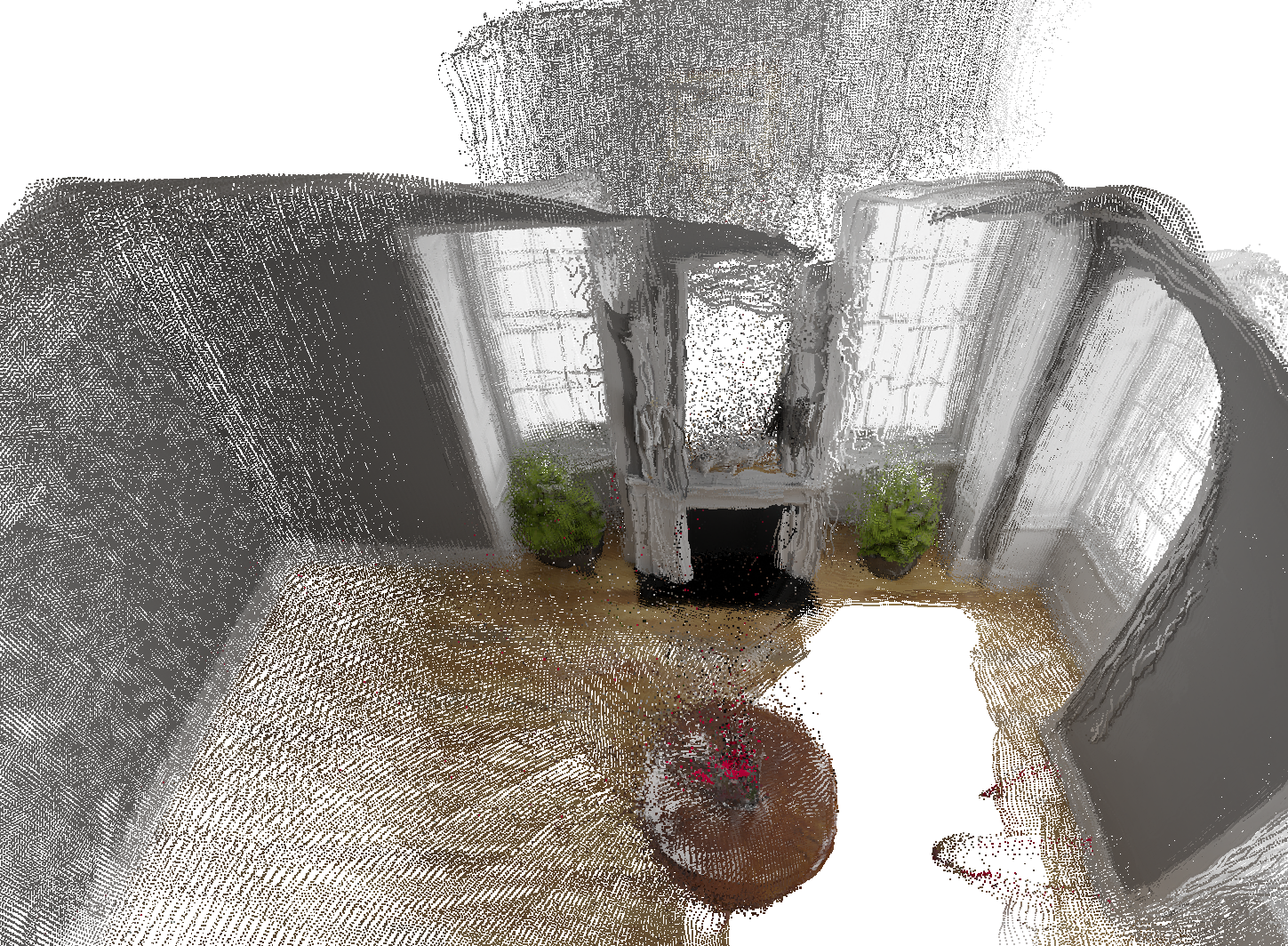} \\

    \end{tabular}
    \vspace{-5pt}
    \caption{\textbf{Qualitative example of outlier scene on NRGBD}. Due to the presence of the mirror, only DUSt3R$^\dagger$ reconstructs the geometry of the mirror and produces fewer floaters. We hypothesize this is due to more synthetic training data used in DUSt3R.}
    \label{fig:gwr_images}
    \vspace{-5pt}
\end{figure}

%% file: Tables/quant_per_scene_nrgbd.tex
\begin{table}[t]
  \centering
  \footnotesize
  \setlength{\tabcolsep}{0.3em}
    \begin{tabularx}{\columnwidth}{ll >{\centering\arraybackslash}X >{\centering\arraybackslash}X >{\centering\arraybackslash}X >{\centering\arraybackslash}X >{\centering\arraybackslash}X >{\centering\arraybackslash}X}
      \toprule
          \multirow{2}{*}{Scene} & \multirow{2}{*}{Method} & \multicolumn{2}{c}{\tt{Acc}$\downarrow$} & \multicolumn{2}{c}{\tt{Comp}$\downarrow$} & \multicolumn{2}{c}{\tt{NC}$\uparrow$} \\ 
      \cmidrule(lr){3-4} \cmidrule(lr){5-6} \cmidrule(lr){7-8}
         & & \tt{Mean} & \tt{Med.} & \tt{Mean} & \tt{Med.} & \tt{Mean} & \tt{Med.} \\
      \midrule
      \multirow{2}{*}{SC} & \textbf{Dust3R$^\dagger$} & \bf 0.0731 & 0.0370 & 0.0296 & 0.0107 & \bf 0.7404 & \bf 0.9018 \\
      
      & \textbf{Ours}             & 0.0740 & \bf 0.0359 & \bf 0.0249 & \bf 0.0106 & 0.7234 & 0.8779\\

      \midrule

      \multirow{2}{*}{CK} & \textbf{Dust3R$^\dagger$} & \bf 0.0553 & \bf 0.0252 & \bf 0.0242 & \bf 0.0113 & \bf 0.8167 & \bf 0.9706 \\
      
      & \textbf{Ours}             & 0.0916 & 0.0356 & 0.0310 & 0.0139 & 0.7811 & 0.9460\\

      \midrule

      \multirow{2}{*}{GWR} & \textbf{Dust3R$^\dagger$} & \bf 0.1097 & \bf 0.0348 & \bf 0.0342 & \bf 0.0170 & \bf 0.8198 & \bf 0.9625 \\
      
      & \textbf{Ours}             & 0.2074 & 0.0646 & 0.0499 & 0.0224 & 0.7628 & 0.9262\\

      \midrule

      \multirow{2}{*}{MA} & \textbf{Dust3R$^\dagger$} & \bf 0.0220 & \bf 0.0154 & \bf 0.0158 & 0.0090 & \bf 0.8126 & \bf 0.9728 \\
      
      & \textbf{Ours}             & 0.0250 & 0.0173 &  0.0160 & \bf 0.0077 & 0.8089 & 0.9677\\

      \midrule

      \multirow{2}{*}{GR} & \textbf{Dust3R$^\dagger$} & \bf 0.0330 & \bf 0.0232 & 0.0554 & \bf 0.0086 & \bf 0.8003 & \bf 0.9534 \\
      
      & \textbf{Ours}             & 0.0486 & 0.0341 & \bf 0.0529 & \bf 0.0101 & 0.7816 & 0.9423\\

      \midrule

      \multirow{2}{*}{Kit.} & \textbf{Dust3R$^\dagger$} & 0.0965 & 0.0438 & 0.0656 & 0.0177 & \bf 0.8157 & \bf 0.9732 \\
      
      & \textbf{Ours}             & \bf 0.0649 & \bf 0.0359 & \bf 0.0333 & \bf 0.0132 & 0.8140 & 0.9683\\

      \midrule

      \multirow{2}{*}{WR} & \textbf{Dust3R$^\dagger$} & \bf 0.0300 & \bf 0.0170 & \bf 0.0119 & \bf 0.0071 & \bf 0.7866 & \bf 0.9352 \\
      
      & \textbf{Ours}             & 0.0426 & 0.0270 & 0.0169 & 0.0081 & 0.7500 & 0.9022\\

      \midrule

      \multirow{2}{*}{BR} & \textbf{Dust3R$^\dagger$} & 0.0476 & \bf 0.0211 & 0.0343 & \bf 0.0061 & 0.7460 & 0.9162 \\
      
      & \textbf{Ours}             & \bf 0.0472 & 0.0215 & \bf 0.0255 & 0.0067 & \bf 0.7613 & \bf 0.9312\\

      \midrule

      \multirow{2}{*}{TG} & \textbf{Dust3R$^\dagger$} & 0.0228 & \bf 0.0084 & 0.0130 & \bf 0.0054 & \bf 0.8838 & \bf 0.9911 \\
      
      & \textbf{Ours}             & \bf 0.0207 & 0.0119 & \bf 0.0120 & 0.0065 & 0.8150 & 0.9719\\

      \midrule

      \multirow{2}{*}{Avg.} & \textbf{Dust3R$^\dagger$} & \bf 0.0544 & \bf 0.0251 & 0.0315 & \bf 0.0103 & \bf 0.8024 & \bf 0.9529 \\
      
      & \textbf{Ours}             & 0.0691 & 0.0315 & \bf 0.0291 & 0.0110 & 0.7775 & 0.9371\\

      \bottomrule
    \end{tabularx}%
    \vspace{-5pt}
    \caption{\textbf{Per-scene results on NRGBD dataset.}}
    \vspace{-8pt}
    \label{tab:per_scene_NRGBD}
\end{table}